\documentclass[11pt]{article}

\usepackage[preprint]{acl}

\usepackage{times}
\usepackage{latexsym}
\usepackage[T1]{fontenc}
\usepackage[utf8]{inputenc}
\usepackage{microtype}
\usepackage{inconsolata}
\usepackage{graphicx}

\usepackage{acro}
\usepackage{amsmath}
\usepackage{amssymb}
\usepackage[british]{babel}
\usepackage{bookmark}
\usepackage{booktabs}
\usepackage{cleveref}
\usepackage{csquotes}
\usepackage[inline]{enumitem}
\usepackage[shortcuts]{extdash}
\usepackage[htt]{hyphenat}
\usepackage{listings}
\usepackage{makecell}
\usepackage{mhchem}
\usepackage{multirow}
\usepackage[all]{nowidow}
\usepackage{pgfplots}
\usepackage{rotating}
\usepackage{siunitx}
\usepackage[most]{tcolorbox}
\usepackage{tikz}
\usepackage{todonotes}
\usepackage{xurl}

\DeclareAcronym{cb}{
  long=closed-book,
  short=CB,
}
\DeclareAcronym{ci}{
  long=confidence interval,
  short=CI,
}
\DeclareAcronym{knn}{
  long=\mbox{\(k\)-nearest} neighbours,
  short=\mbox{\(k\)-NN},
}
\DeclareAcronym{lerp}{
  long=linear interpolation,
  short=\textsc{lerp},
}
\DeclareAcronym{lora}{
  long=low-rank adaptation,
  short=LoRA,
}
\DeclareAcronym{nmi}{
  long=normalized mutual information,
  short=NMI,
}
\DeclareAcronym{ob}{
  long=open-book,
  short=OB,
}
\DeclareAcronym{qa}{
  long=question-answering,
  short=QA,
}
\DeclareAcronym{rag}{
  long=retrieval-augmented generation,
  short=RAG,
}
\DeclareAcronym{squad}{
  long=Stanford Question Answering Dataset,
  short=SQuAD,
}
\DeclareAcronym{svd}{
  long=singular value decomposition,
  short=SVD,
}
\DeclareAcronym{ties}{
  long={trim, elect sign, and merge},
  short=\textsc{ties},
}

\pgfplotsset{compat=1.18}
\usepgfplotslibrary{groupplots}

\setuptodonotes{size=\small}

\newtcblisting{prompt}{
  breakable,
  listing only,
  listing options={
    basicstyle=\ttfamily,
    breakatwhitespace=true,
    breakautoindent=false,
    breakindent=0pt,
    breaklines=true,
    columns=fullflexible,
    extendedchars=true,
    inputencoding=utf8,
    keepspaces=true,
    literate=
    {–}{{--}}1
    {—}{{---}}1
    {…}{{\ldots}}1
    {’}{{'}}1
    {‘}{{`}}1
    {”}{{''}}1
    {“}{{``}}1
    {±}{{$\pm$}}1
    {≤}{{$\le$}}1
    {≥}{{$\ge$}}1,
    postbreak=\mbox{\textcolor{gray}{$\hookrightarrow$}\space},
  }
}

\definecolor{my_blue_text}{HTML}{086179}
\definecolor{my_blue}{HTML}{0B81A2}
\definecolor{my_darkblue_text}{HTML}{082A54}
\definecolor{my_darkblue}{HTML}{082A54}
\definecolor{my_darkred_text}{HTML}{9D2C00}
\definecolor{my_darkred}{HTML}{9D2C00}
\definecolor{my_gold_text}{HTML}{76520C}
\definecolor{my_gold}{HTML}{C58915}
\definecolor{my_gray_text}{HTML}{595959}
\definecolor{my_gray}{HTML}{949494}
\definecolor{my_green_text}{HTML}{1F6700}
\definecolor{my_green}{HTML}{33AB00}
\definecolor{my_purple_text}{HTML}{78438D}
\definecolor{my_purple}{HTML}{7E4794}
\definecolor{my_red_text}{HTML}{AD1E20}
\definecolor{my_red}{HTML}{E25759}
\definecolor{my_teal_text}{HTML}{33615A}
\definecolor{my_teal}{HTML}{55A296}

\newcommand{\adddatapointci}[9]{
  \addplot[
    color=#1,
    error bars/x dir=both,
    error bars/x explicit,
    error bars/y dir=both,
    error bars/y explicit,
    mark=*,
    mark options={fill=#1},
    only marks,
  ] coordinates {
    (#2, #3) += (#5, #7) -= (#4, #6)
  };
  \node[
    color=#1_text,
    #8,
  ] at (axis cs: #2, #3) {#9};
}

\newcommand{\estci}[3]{\num{#1} [\num{#2}, \num{#3}]}

\newcommand{\sqrtlerp}{\ensuremath{\surd}\textsc{lerp}}

\NewDocumentCommand{\ci}{s g g}{%
  \ensuremath{%
    \mathrm{CI}_{\qty{95}{\percent}}%
    \IfValueT{#2}{ %
      = %
      \IfBooleanTF{#1}%
      {[\qty{#2}{\percent}, \qty{#3}{\percent}]}%
      {[\num{#2}, \num{#3}]}%
    }%
  }%
}

\title{
  Mostly Reading: What Open-Book Scores Measure \\
  in Document-Level LoRA Adapters
}

\author{
  \textbf{Martino M.\ L.\ Pulici\textsuperscript{1,2}},
  \textbf{Cuong Xuan Chu\textsuperscript{1}},
  \textbf{Evgeny Kharlamov\textsuperscript{1}},
  \textbf{Volker Tresp\textsuperscript{2,3}}
  \\
  \\
  \textsuperscript{1}Bosch Center for Artificial Intelligence, Germany
  \\
  \textsuperscript{2}LMU Munich, Germany
  \\
  \textsuperscript{3}Munich Center for Machine Learning, Germany
  \\
  \small{
    \textbf{Correspondence:} \href{mailto:martino.pulici@de.bosch.com}{martino.pulici@de.bosch.com}
  }
}

\begin{document}

\maketitle

\begin{abstract}
  Single-document LoRA adapters have been proposed as units of knowledge to be merged and retrieved; yet open-book scoring, with the document in the prompt, can credit a document's own adapter (its specialist) for reading that document as readily as for holding it.
  Training one adapter for each of \num{150} Wikipedia articles on Llama-3.1-8B-Instruct, we find that open-book scores mostly measure a reading behaviour every adapter shares: a random adapter trained on another document recovers \qty{82}{\percent} of the specialist's open-book advantage over the base model but only \qty{2}{\percent} of its advantage on a cloze probe that asks for the same document's facts without providing the source text, a share we cannot distinguish from zero.
  Content does reach the weights, since the specialists beat the base model in cloze recall by \num{0.094}, and the geometry of their updates follows the documents they read; yet, in a test of \num{30} queries, adapters chosen by weight-space proximity or text embedding are not distinguishable from random ones on either probe, although that test cannot exclude an advantage of about four fifths of what a query's own adapter gains over a random one.
  At \num{16} documents or more, correctly scaled merges recover \qtyrange{87}{89}{\percent} of the specialists' open-book advantage but only \qtyrange{12}{16}{\percent} of their cloze advantage, with intervals that run from below zero to at most \qty{32}{\percent}, against \qty{94}{\percent} for joint training.
  We also report four silent errors, each of which produced a clean, wrong~result.
\end{abstract}

\section{Introduction}

\Acl{lora} (\acs{lora}; \citealp{hu2022lora}) makes fine-tuning cheap enough that adapters can be imagined as a storage medium, one per document, kept in a bank and merged \citep{ilharco2023editing,yadav2023ties} or retrieved \citep{su2025prag,back2026understanding} as a task requires.
What such a bank is worth depends on what its adapters hold; yet scoring them \ac{ob}, with the document in the prompt as in the variant of parametric \ac{rag} that \citet{tang2025understanding} find strongest, cannot tell an adapter that holds the document's content from one that has learned to extract answers from whatever text it is given: we find that it mostly measures the second.
To separate the two, on Llama-3.1-8B-Instruct \citep{llamateam2024llama}, we change which adapter answers a fixed document in place of its own specialist, and we employ a cloze probe that states each held-out fact as a sentence with one blank\---without the document and in a surface form that no adapter was trained on (\cref{sec:setup}).

Holding each document's training questions fixed and varying the text trained beside them (\cref{sec:conditions}), we find that an adapter trained on the questions without the text comes second of five training conditions when scored \ac{ob}, yet on cloze it recalls \num{0.132} less of the document than the base model does; a cloze gain clear of zero appears exactly where an adapter read the scored document's full text, and the specialists' gain over the base model is \num{0.094}.
The weights record the same variable: distances between updates sort adapters by domain (\cref{sec:semantic-structure}) and, when an adapter's questions and text come from different documents, follow the text rather than the questions (\cref{sec:cs}).

\Ac{ob} scores follow something else.
An adapter applied to another article of its own domain recovers \qty{88}{\percent} of the \ac{ob} advantage that the specialist holds over the base model (\cref{sec:transfer}), and pooled over collections of \num{16} documents or more, a random adapter from outside the collection recovers \qty{82}{\percent} of the specialist's \ac{ob} advantage and \qty{2}{\percent} of its cloze advantage, with a \ac{ci} that includes zero (\cref{sec:decomposition}).
In a bank that never holds the query's own adapter, an adapter chosen by weight-space proximity is not distinguishable from a random one on either probe, nor is one chosen by text embedding, although with \num{30} queries the cloze \ac{ci} still admits an advantage of about four fifths of what the query's own adapter gains over a random one (\cref{sec:retrieval}).

Composition shows the same split (\cref{sec:composition}): at four documents or more, merges keep \numrange{0.93}{1.01} of the specialists' \ac{ob} score against \numrange{0.88}{0.97} for random, but at \num{16} documents or more they recover \qtyrange{12}{16}{\percent} of the cloze advantage, each \ac{ci} including zero, where joint training recovers \qty{94}{\percent}.
\Ac{ob} scores thus overstate how much of a document's content survives merging or substitution.

Each of four silent errors produced clean results we were prepared to report (\cref{sec:validity}):
\begin{enumerate*}[label= (\emph{\alph*})]
  \item a training-length default that reversed a comparison between training conditions;
  \item a repetition penalty left in our evaluation code;
  \item a scaling error in our merge, which looked like a cost of merging; and
  \item an evaluation precision unlike the one the adapters were trained on.
\end{enumerate*}

\section{Experimental setup}\label{sec:setup}

\paragraph{Corpus and training.}
We use \num{150} Wikipedia articles in English \citep{wikimediafoundation2026wikipedia}, \num{25} from each of biology, computer science, economics, history, law, and medicine.
For each article Claude Sonnet 4.6 \citep{anthropic2026introducing} generated \num{12} \ac{qa} pairs, each validated by string match against the source text; eight are used for training, and four (three factual questions and one true/false item) are held out for evaluation.
The base model is Llama-3.1-8B-Instruct \citep{llamateam2024llama,metallama2024llama31}, loaded in \num{4}-bit NF4 with bf16 computation as in QLoRA \citep{dettmers2023qlora}; adapters of rank \num{8} on all seven projection matrices train for \num{500} steps with next-token cross-entropy on every token of the training string (instruction, document, question, and answer), capped at \num{2048} tokens, which fits every training example.
The geometry of \cref{sec:geometry} is replicated with the same hyper-parameters on three other models; \cref{app:software} gives the rest of the configuration.

\paragraph{Training-text conditions.}
Each document's \ac{qa} pairs are held fixed, while the text trained beside them (generated by Claude Sonnet 4.6 from the prompts of \cref{app:prompts}) varies, resulting in five conditions:
\begin{enumerate*}[label= (\emph{\alph*})]
  \item the \textbf{natural} article;
  \item a \textbf{style-normalized} rewrite into neutral prose;
  \item a \textbf{synthetic} fill-in-the-blank template that keeps six extracted facts;
  \item \textbf{no context}, which omits the document text; and
  \item \textbf{swapped context}, including another article's prose beside this article's questions.
\end{enumerate*}
When no condition is explicitly named, the adapter is trained on the natural article.

\paragraph{Evaluation.}
Every evaluation loads the base model in \num{4}-bit NF4 and applies each adapter as a \ac{lora} module.
\Ac{ob} evaluation supplies the document with the training-time instruction template (\cref{app:prompt-instruction}) without a chat template, at a window of \num{2048} tokens that cuts no prompt, while \ac{cb} evaluation withholds it; answers are generated greedily, up to \num{128} new tokens and with no repetition penalty, and scored by \ac{squad} token-level \(F_1\) \citep{rajpurkar2016squad} over all four held-out items; \cref{app:heldout} checks the metric against answer length.
Semantic distance between documents is one minus the cosine similarity of their all-mpnet-base-v2 sentence embeddings \citep{reimers2019sentencebert,sentencetransformers2021allmpnetbasev2}.

\paragraph{The cloze probe.}
The cloze probe rewrites each held-out \ac{qa} pair as a declarative sentence with one blank that the answer fills (\cref{app:prompt-qa}) and presents it without the document, so that filling it requires the fact from the weights in a surface form that no adapter was trained on.
The three factual items are scored by recall of the reference fill (an empty prediction scoring zero), and the true/false item is reported beside them.
Unlike the \ac{qa} evaluations the cloze harness wraps each prompt in the model's chat template, which no adapter saw in training but which enters every score and every reference alike; rerun in the training format on \num{35} documents, the chat wrapper raises the specialists' cloze gain over the base model by \num{+0.057}, a change whose \ac{ci}, \([\num{-0.069}, \num{+0.183}]\), includes zero (\cref{app:template}).

\paragraph{Confidence intervals.}
\Acp{ci} are \qty{95}{\percent} percentile intervals from \num{10000} bootstrap resamples, paired across methods; the resampled unit is the document, except for composition contrasts (for which it is the collection), retrieval (the query), and transfer (the source--target~pair).

\paragraph{Composition protocols.}
For a collection of \(N\) documents we compare five ways of supplying one adapter to answer all of the questions, the first three of which are merging methods:
\begin{enumerate*}[label= (\emph{\alph*})]
  \item \textbf{\sqrtlerp{}}, a \acl{lerp}\acuse{lerp} of the stored factors at square-root weight forming \((N^{-1/2}\sum_i \mathbf{B}_i) (N^{-1/2}\sum_j \mathbf{A}_j)\), in which each adapter's own update is weighted by \(1 / N\) (\cref{sec:mergespace});
  \item \textbf{update merging}, which averages the updates \(\mathbf{B}_i \mathbf{A}_i\) and projects the average back to rank \num{8} by truncated \acl{svd} (\acs{svd}; \cref{app:operators});
  \item \acl{ties} (\textbf{\acs{ties}}; \citealp{yadav2023ties}), applied to the \(\mathbf{A}\) and \(\mathbf{B}\) factors separately (\cref{app:operators});
  \item \textbf{joint} training, which fits one adapter to the union of the collection's training examples for the same \num{500} steps and should be viewed as a retraining reference rather than a bound; and
  \item \textbf{random}, applying one adapter trained on a document outside the collection and averaging its scores over five independent draws because single values vary widely (\cref{app:rand}).
\end{enumerate*}
\textbf{Retention} is a method's score on a document's held-out items divided by the score of the document's own specialist adapter.
Collections are drawn at \(N \in \{2, 4, 8, 16, 25, 32\}\) under a \textbf{within-domain} regime and a \textbf{cross-domain} one spread over all six domains, five per cell with seeds shared across methods so that every comparison is paired, \num{56} collections in all; within-domain at \(N = 25\) uses the six domains themselves, so in that setting every random adapter necessarily comes from another domain.

\section{Content and geometry}\label{sec:geometry}

\subsection{Training text and encoding}\label{sec:conditions}

\Cref{tab:conditions} scores the training-text conditions of \cref{sec:setup}, which fix each document's \ac{qa} pairs and vary only the text trained beside them; cloze figures are paired differences from the base model over the \num{150} documents, since the probe's absolute scale is not comparable to that of the \ac{qa} evaluations, and \cref{app:conditions} gives every \ac{ci}.
For the \acs{ob} evaluation the conditions read different texts, listed in \cref{app:conditions}; swapped context scored on its source, in particular, reads the donor document with the source's questions, so its context holds none of the answers.

\begin{table}
  \centering

  \begin{tabular}{l *{2}{S[table-format=1.3]} S[table-format=+1.3]}
    \toprule
                                &\multicolumn{2}{c}{Mean \(F_1\)} & \\
    \cmidrule(lr){2-3}
    \multirow{-2}{*}{Condition} & {\Acs{ob}}                      & {\Acs{cb}} & {\multirow{-2}{*}{\makecell{\(\Delta\) cloze\\recall}}} \\
    \midrule
    Base                        & 0.504                           & 0.320      & {---} \\
    Natural                     & 0.822                           & 0.399      & +0.094 \\
    Style-normalized            & 0.708                           & 0.419      & +0.097 \\
    Synthetic                   & 0.649                           & 0.443      & +0.034 \\
    No context                  & 0.773                           & 0.426      & -0.132 \\
    Swapped (source)            & 0.424                           & 0.456      & -0.028 \\
    Swapped (donor)             & 0.686                           & 0.428      & +0.056 \\
    \bottomrule
  \end{tabular}

  \caption{
    Token \(F_1\) over all \num{150} documents (\acs{ob} and \acs{cb}) and cloze recall on each document's three factual items as a paired difference from the base model.
    All \acsp{ci} exclude zero except for synthetic and swapped context on source; each \acs{ob} condition reads the text listed in \cref{app:conditions}, with swapped context on source reading its donor's text (which holds none of its answers).
  }\label{tab:conditions}
\end{table}

\Ac{cb} \(F_1\) and cloze recall rank the conditions in nearly opposite orders.
The three conditions that lead \ac{cb} are those that read least of the scored document's content (swapped context scored on its source at \num{0.456}, synthetic at \num{0.443}, and no context at \num{0.426}, against \num{0.399} for the natural adapter and \num{0.320} for the base model), yet on cloze the swapped and synthetic adapters are not distinguishable from the base model, and the no-context adapters fall below it by \num{0.132} (\ci{0.087}{0.176}), further than the natural adapter rises above it (\num{+0.094}).
A cloze gain clear of zero appears exactly where an adapter read the scored document's full content: the natural adapter, its style-normalized rewrite, and the swapped adapter scored on the document it read.
Natural and style-normalized are basically equal (\num{-0.003} with \ci{-0.042}{+0.037}), whereas the style-normalized adapters exceed the synthetic ones by \num{0.063} with a \ac{ci} clear of zero; the synthetic condition sees different facts during training and evaluation, so its missing gain may reflect coverage as much as format.
We therefore read \ac{cb} scores as rewarding answer-shaped output in the format an adapter trained on, most readily supplied by the conditions without usable prose, while cloze, which states each fact in a surface form that no condition trained on, is the probe that tracks content.

\Ac{ob} scores follow neither ordering: the no-context adapter, which read no document text at all, comes second of the five conditions at \num{0.773}, \num{0.049} (\ci{0.025}{0.074}) behind the natural adapter and ahead of both rewrites, and every neighbouring pair in that column differs by a paired \ac{ci} that excludes zero. Therefore, an adapter can score close to a specialist \ac{ob} while recalling less of the document than the unadapted model does.

\subsection{Orthogonality and semantic structure}\label{sec:semantic-structure}

Merging acts on the weight updates themselves, whose cosine (averaged over all \num{224} layer--module combinations) has a mean of \num{0.0066} across the \num{11175} pairs, every pair lying between \num{0.004} and \num{0.013}.
Two adapters driven by identical gradients through independently initialized \(\mathbf{A}\) matrices would give about \(r / d_{\mathrm{in}}\) (i.e.\ \num{0.0018} averaged over the \num{224} combinations), so a small cosine cannot by itself show that adapters hold different things; the observed mean is nearly four times that reference, which indicates that the \(\mathbf{A}\) matrices move toward input directions the adapters share (\cref{app:math}), although for merging the updates stay nearly orthogonal layer by layer.

Pairwise Frobenius distances between adapter updates correlate with sentence-embedding distances between the corresponding documents on every architecture we train.
Before these runs we had fixed a Spearman \(\rho\) of \num{0.250} as the threshold a single layer would need to be selected; no single layer reached it, whereas with every layer aggregated three of the four architectures exceed it, with Llama-3.1-8B-Instruct falling just short at \num{0.242}.
The correlation concentrates in the final third of each network: on Llama-3.1-8B-Instruct selecting the eight highest-correlating layers on half the adapters and evaluating them on the other half gives a held-out \(\rho\) of \num{0.296} (\ci{0.204}{0.356}) against \num{0.318} when the same adapters both select and score the layers.
Over all layers the fraction of an adapter's six nearest neighbours sharing its domain is \num{0.708} against a chance rate of \num{0.161}, and clustering separates sub-domains that the six labels do not encode (\cref{app:geometry} gives every architecture, the selection procedure, and the clusters): if proximity in this space fails to pick out adapters that carry a document's content, an uninformative geometry is not the likely explanation.

\subsection{Context swap}\label{sec:cs}

Because the document an adapter reads and the questions it answers come from the same source in every condition but one, the correlation above cannot say which of the two is recorded by the geometry: the swapped-context condition separates them.
The geometry follows the donor, correlating with its semantics at \(\rho = \num{+0.217}\) against \num{+0.033} for the source, and not because the swapped adapters fail to move, since their updates travel as far as the natural condition's (mean pairwise distance of \num{21.09} against \num{21.10}).
Cloze recall points the same way: the swapped adapters exceed the base model by \num{0.056} on the document they read, whereas on their source they are not distinguishable from it.
Paired on the \num{99} distinct donor documents, a swapped adapter reaches an \ac{ob} \(F_1\) of \num{0.685}, against \num{0.812} for the document's natural adapter and \num{0.503} for the base model: the swapped adapter recovers \qty{59}{\percent} (\ci*{48}{69}) of the natural adapter's advantage, with question supervision supplying the rest (\cref{app:swap}).

\section{Behaviour and content}\label{sec:behaviour}

\Ac{ob} scoring supplies the document in the prompt, as does the variant of parametric \ac{rag} that also places the retrieved passages in context, which \citet{tang2025understanding} find the strongest: an adapter can therefore be credited for reading a supplied text well as readily as for holding what it read, and we separate the two by fixing the scored document and changing which adapter answers it.
We express each result as a share of the specialists' advantage over the base model on the same documents: \((\bar{s} - \bar{s}_b) / (\bar{s}_s - \bar{s}_b)\), where \(\bar{s}\), \(\bar{s}_s\), and \(\bar{s}_b\) are the mean scores of the method, of the documents' own specialists, and of the base model respectively.
We give shares as percentages and retentions, which divide scores rather than advantages, as decimals\---this way the two kinds of ratio cannot be mistaken for one another.

\subsection{Transfer to unseen documents}\label{sec:transfer}

When applied \ac{ob} to a different article from its own domain, over \num{120} source--target pairs, an adapter raises the base model's score on the target by \num{+0.273} (\ci{+0.245}{+0.301}), while each specialist adds \num{+0.309} (\ci{+0.283}{+0.335}) on its own article.
A foreign adapter therefore recovers \qty{88}{\percent} (\ci*{77}{101}) of the specialist's advantage;
the \ac{ci} reaches \qty{100}{\percent}, so the part of that advantage which does not transfer\---the only part \ac{ob} scoring could attribute to the document itself\---cannot be distinguished from zero.
No-context adapters, trained on the target's questions without its text, gain \num{+0.242} (\ci{+0.214}{+0.270}), or \qty{78}{\percent} (\ci*{68}{90}) of the specialists' advantage.
What an adapter carries to another document is therefore mostly a disposition to extract answers from whatever context it is given, which applies to any document; \cref{sec:decomposition} asks how much of it survives once the document is withheld.

\subsection{Random adapter decomposition}\label{sec:decomposition}

\begin{figure*}[t]
  \centering

  \begin{tikzpicture}
    \begin{axis}[
        axis lines=left,
        clip mode=individual,
        grid=major,
        height=0.414\textwidth,
        width=\textwidth,
        xlabel={Share of the open-book advantage (\%)},
        xmax=1.075,
        xmin=0.475,
        xticklabel={\pgfmathparse{100*\tick}\pgfmathprintnumber[
            fixed,
            precision=0,
        ]{\pgfmathresult}},
        ylabel={Share of the cloze advantage (\%)},
        ylabel absolute,
        ymax=1.5,
        ymin=-1.5,
        yticklabel={\pgfmathparse{100*\tick}\pgfmathprintnumber[
            fixed,
            precision=0,
        ]{\pgfmathresult}},
      ]
      \addplot[
        color=black,
        mark=none,
        thin,
      ] coordinates {
        (0.475,0)
        (1.075,0)
      };
      \addplot[
        color=black,
        dotted,
        mark=none,
        thick,
      ] coordinates {
        (0.475,1)
        (1.075,1)
      };
      \addplot[
        color=black,
        dotted,
        mark=none,
        thick,
      ] coordinates {
        (1,-1.5)
        (1,1.5)
      };
      \addplot[
        color=my_gray,
        dashed,
        domain=0.475:1.075,
        samples=2,
        thick,
      ] {x};
      \adddatapointci{my_gold}{0.887}{0.122}{0.067}{0.075}{0.186}{0.178}{above right}{\(\surd\)\textsc{Lerp}}
      \adddatapointci{my_teal}{0.872}{0.125}{0.070}{0.078}{0.210}{0.186}{above left}{Update merging}
      \adddatapointci{my_blue}{0.882}{0.156}{0.068}{0.076}{0.169}{0.168}{above, yshift=3mm}{\textsc{Ties}}
      \adddatapointci{my_red}{0.992}{0.938}{0.063}{0.068}{0.239}{0.456}{below left}{Joint}
      \adddatapointci{my_darkred}{0.817}{0.024}{0.051}{0.058}{0.219}{0.144}{below left}{Random}
      \adddatapointci{my_purple}{0.846}{-1.394}{0.075}{0.075}{1.363}{0.651}{above right}{No context}
      \adddatapointci{my_green}{0.589}{0.568}{0.111}{0.103}{0.514}{0.421}{below right}{Swapped context (donor)}
    \end{axis}
  \end{tikzpicture}

  \caption{
    Share of the specialists' advantage over the base model that each method recovers, \acs{ob} on all four held-out items against cloze on the factual items, at \(N \ge 16\).
    Each point is a ratio of means with its \qty{95}{\percent} \ac{ci} from \num{10000} bootstrap resamples over documents; the dotted lines mark full recovery on each probe, and the dashed diagonal indicates equal recovery on both.
    The shares of no context and swapped context are divided by the natural adapter's advantage over the base model on the documents each is scored on\---all \num{150} for no context, and its \num{99} donor documents for swapped context; the no-context adapter's cloze \ac{ci}, \([\qty{-276}{\percent}, \qty{-74}{\percent}]\), runs below the axis.
  }\label{fig:thesis}
\end{figure*}
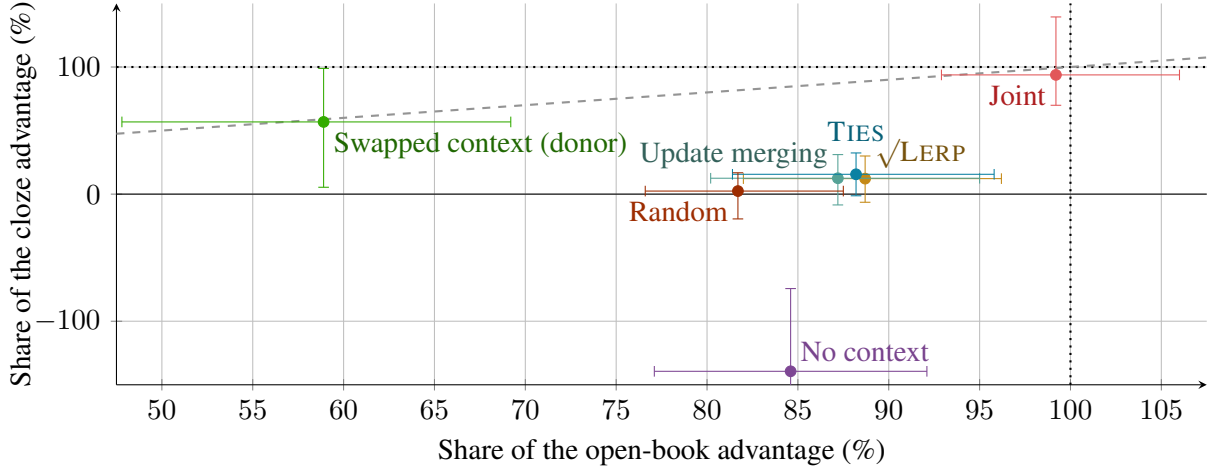

Pooled over the \num{26} collections of the composition grid that hold \num{16} documents or more (\num{595} document rows), the random adapter recovers \qty{82}{\percent} (\ci*{77}{87}) of the specialists' \ac{ob} advantage over the base model and \qty{2}{\percent} (\ci*{-19}{17}) of their cloze advantage, a share not distinguishable from zero (\cref{fig:thesis}; by regime in \cref{app:rand}).

The cloze share carries a wider \ac{ci} because its margin is smaller: over these rows the specialists lead the base model by \num{+0.330} \ac{ob} and by \num{+0.105} in cloze recall, where the base model already reaches \num{0.81} of their score, against \num{0.61} of their \ac{ob} score over the \num{150} documents; both shares therefore divide what adapters add to a model that already does much of the task.
The probe is not blind to recovery: joint training, scored on \num{25} of these collections, recovers \qty{94}{\percent} (\ci*{70}{139}) of the cloze advantage and \qty{99}{\percent} (\ci*{93}{106}) of the \ac{ob} one (\cref{sec:cloze-grid}).

\subsection{Retrieval by weight-space proximity}\label{sec:retrieval}

Since an adapter's weights record which document it read (\cref{sec:cs}), we ask whether a bank of \num{120} specialists, searched by weight-space proximity, yields an adapter that carries the content of each of \num{30} other documents (five per domain) whose own adapters the bank never holds.
For each query we compare four settings:
\begin{enumerate*}[label= (\emph{\alph*})]
  \item the bank adapter nearest to the query's own using the all-layer Frobenius distance;
  \item the one whose document lies nearest in sentence-embedding distance;
  \item five averaged random adapters from the whole bank; and
  \item five averaged ones from the query's domain.
\end{enumerate*}
Before computing any result, we fix the primary contrast: weight-space selection against random adapters from the whole bank on the cloze factual items. The chosen criterion is that weight space counts as better only if that contrast's \ac{ci} lay above~zero.

An adapter chosen by weight-space proximity is not distinguishable from a random one on either probe (\cref{tab:retrieval}), and neither is one chosen by text embedding: on cloze the primary contrast is \num{+0.020} (\ci{-0.038}{+0.075}) and text embedding against random gives \num{-0.001} (\ci{-0.060}{+0.061}), while \ac{ob} the same weight-space selections score \num{-0.024} (\ci{-0.061}{+0.013}) against random.
These \acp{ci} are wide by design: on cloze the query's own adapter scores \num{0.091} above a random one (\num{0.507} against \num{0.415}), and the primary contrast's \ac{ci} reaches \num{+0.075}, about four fifths of that margin, so the run finds no advantage for weight-space selection but cannot rule out one nearly as large as that of the query's own adapter.
Over these \num{30} documents even the own adapter is not distinguishable from the base model on cloze (\num{+0.085}), although over all \num{150} its gain is clear of zero; a share of that margin would have an undetermined denominator, so we report retrieval as differences in recall, and \cref{fig:thesis} carries no retrieval point.
With one adapter selected, every arm that uses a foreign adapter lies within \num{0.03} of the base model on cloze, yet recovers \qtyrange{70}{80}{\percent} of the own adapter's \ac{ob} advantage over it.
Weight space places its choice in the query's domain for \(\num{26} / \num{30}\) queries, where random choice would give \(\num{5} / \num{30}\), since each domain holds a sixth of the bank: geometry that sorts adapters by domain did not, in our bank, pick out adapters that help with a document's content.
\Cref{app:retrieval} gives the criterion's timeline and all \num{24} declared~contrasts.

\begin{table}
  \centering

  \begin{tabular}{l *{2}{S[table-format=1.3]}}
    \toprule
    Adapter applied & {\Acs{ob} \(F_1\)} & {Cloze recall} \\
    \midrule
    None            & 0.510              & 0.422 \\
    Weight space    & 0.726              & 0.435 \\
    Text embedding  & 0.757              & 0.414 \\
    Random (bank)   & 0.750              & 0.415 \\
    Random (domain) & 0.757              & 0.395 \\
    \midrule
    Query's own     & 0.818              & 0.507 \\
    \bottomrule
  \end{tabular}

  \caption{
    \acs{ob} token \(F_1\) on all four held-out items and cloze recall on the three factual items.
    One adapter is selected for each of \num{30} query documents from a bank of \num{120} that never holds the query's own.
    The random rows average five draws, from the whole bank and from the query's domain respectively.
  }\label{tab:retrieval}
\end{table}

\section{Composition}\label{sec:composition}

Since \ac{ob} scores mostly credit a behaviour that every adapter shares, we score the composition grid of \cref{sec:setup} on both probes.

\subsection{Merges on open book}\label{sec:merges-ob}

At rank \num{8} the three merges keep between \num{0.93} and \num{1.01} of the specialists' \ac{ob} score over the nine cells with \(N \ge 4\) (\cref{tab:matchedrank}): \sqrtlerp{} preserves \numrange{0.95}{1.01}, update merging \numrange{0.93}{0.99}, and \ac{ties} \numrange{0.94}{1.00}\---against \numrange{0.88}{0.97} for random.
Paired over the collections of each cell, \sqrtlerp{} lies above random in all nine cells, by \num{0.014} to \num{0.117}, with the \acp{ci} excluding zero in six; update merging's \acp{ci} exclude zero in six cells as well, and those of \ac{ties} in seven (\cref{app:mergespace}).
We do not test the merges against parity with the specialists (update merging lies below it in all nine cells), because a retention near one says little: the reference is the random adapter, which already recovers about four fifths of the specialists' \ac{ob} advantage over the base model (\cref{sec:decomposition}).

\begin{table}
  \centering

  \begin{tabular}{l S[table-format=2.0] *{4}{S[table-format=1.3]}}
    \toprule
    Reg.                   & {\(N\)} & {\(\surd\)\textsc{Lerp}} & {UM}  & {\textsc{Ties}} & {Rand.} \\
    \midrule
    \multirow{5}{*}{Cross} & 4       & 0.996                    & 0.961 & 0.980           & 0.878 \\
                           & 8       & 1.011                    & 0.994 & 1.004           & 0.963 \\
                           & 16      & 0.948                    & 0.932 & 0.942           & 0.933 \\
                           & 25      & 0.978                    & 0.961 & 0.973           & 0.918 \\
                           & 32      & 0.954                    & 0.960 & 0.968           & 0.941 \\
    \midrule
    \multirow{4}{*}{With.} & 4       & 0.982                    & 0.948 & 0.990           & 0.917 \\
                           & 8       & 0.986                    & 0.976 & 0.985           & 0.944 \\
                           & 16      & 0.981                    & 0.989 & 0.972           & 0.963 \\
                           & 25      & 0.996                    & 0.983 & 0.987           & 0.968 \\
    \bottomrule
  \end{tabular}

  \caption{
    \Acs{ob} retention at rank \num{8} in the nine cells with \(N \ge 4\).
    Each document's \(F_1\) under the method is divided by that of its own specialist, averaged over documents, and then over the cell's sets; \cref{app:mergespace} gives the scaling error, the controls, and every paired \ac{ci}.
    \enquote{Reg.} stands for regime, \enquote{UM} for update merging, \enquote{rand.} for random, and \enquote{with.} for within; random averages five out-of-collection adapters per set.
  }\label{tab:matchedrank}
\end{table}

\subsection{Scale and cross terms}\label{sec:mergespace}

Scaling each summed factor by \(N^{-1/2}\), as \sqrtlerp{} does, gives \(N^{-1} \sum_{i} \mathbf{B}_i \mathbf{A}_i + N^{-1} \sum_{i \neq j} \mathbf{B}_i \mathbf{A}_j\) (\cref{app:operators}): so each adapter's own update is weighted by \(1 / N\), as in update merging, together with \(N (N - 1)\) cross terms that pair one adapter's output directions with another's input directions and belong to no trained adapter.
Averaging each factor at \(1 / N\) instead, as our first \ac{lerp} runs did through the error described in \cref{sec:validity}, divides both sums by a further \(N\), which against \sqrtlerp{} lowers \ac{ob} retention by \numrange{0.176}{0.348}, every \ac{ci} excluding zero.

Update merging matches \sqrtlerp{} in own-term weight and parameter count, truncating where \sqrtlerp{} adds the cross terms: \ac{ob} it scores between \num{0.035} below \sqrtlerp{} and \num{0.008} above it, and the four of nine \acp{ci} that exclude zero all favour \sqrtlerp{}; on cloze, pooled over the \num{56} collections, \sqrtlerp{} scores \num{+0.009} (\ci{-0.020}{+0.038}) above update merging, three of the four cells whose \acp{ci} exclude zero favouring \sqrtlerp{} and one favouring update merging.
The gap that \citet{tang2025lora} attribute to the cross terms also carries the \(1 / N^2\) own-term scale of their factor merge (\cref{app:mergespace}); at matched own-term weight the cross terms cost about ten points on the eight vision tasks to which \citet{zheng2025decouple} apply them, against at most \num{0.035} of \ac{ob} retention on our documents.
Truncating to rank \num{8} costs at most \num{0.034} of retention, \ac{ties} differs from update merging by at most \num{0.041} in either direction, and a tuned merging coefficient adds nothing over \(1 / N\) (\cref{app:mergespace}).

\subsection{Joint training}\label{sec:joint}

Pooled over the collections of \num{16} documents or more, joint training exceeds \sqrtlerp{}, update merging, and \ac{ties} by \numrange{0.038}{0.045} of the specialists' \ac{ob} score and by \numrange{0.172}{0.183} of their cloze recall, no \ac{ci}'s lower end falling below \num{+0.023} \ac{ob} or \num{+0.114} on cloze, although per cell the ordering is less uniform (\cref{app:grid}).
At cross-domain with \(N = 32\) joint training is not distinguishable from parity with the specialists, while its lead over random excludes zero: a \num{16}-fold training budget changes retention by less than \num{0.02} (\cref{app:grid}).

\subsection{Merges on cloze}\label{sec:cloze-grid}

\begin{figure*}[t]
  \centering

  \begin{tikzpicture}
    \begin{axis}[
        clip=false,
        height=0.414\textwidth,
        legend style={
          /tikz/every even column/.style={column sep=15mm},
          legend columns=-1,
        },
        legend to name={clozegridlegend},
        width=\textwidth,
        xmax=6.5,
        xmin=0.5,
        xtick={
          1,
          ...,
          6
        },
        xtick style={draw=none},
        xticklabel style={align=center},
        xticklabels={
          16,
          25,
          32,
          16,
          25,
          All \(N\),
        },
        ylabel={Advantage over random},
        ylabel absolute,
        ymajorgrids,
        ymax=0.4,
        ymin=-0.1,
        ytick style={draw=none},
        ytick={
          -0.1,
          0,
          0.1,
          0.2,
          0.3,
          0.4
        },
        yticklabel style={
          /pgf/number format/fixed,
          /pgf/number format/precision=1,
        },
      ]
      \path[every node/.style={anchor=near xticklabel}]{
        (xticklabel cs:0.25)   node {Cross}
        (xticklabel cs:0.6667) node {Within}
        (xticklabel cs:0.9167) node {Pool}
      };
      \addplot[
        black,
        forget plot,
        thin,
      ] coordinates {
        (0.5, 0)
        (6.5, 0)
      };
      \addplot[
        black,
        forget plot,
        loosely dotted,
        thin,
      ] coordinates {
        (1.5, -0.15)
        (1.5, 0.4)
      };
      \addplot[
        black,
        forget plot,
        loosely dotted,
        thin,
      ] coordinates {
        (2.5, -0.15)
        (2.5, 0.4)
      };
      \addplot[
        black,
        forget plot,
        densely dotted,
        thin,
      ] coordinates {
        (3.5, -0.2)
        (3.5, 0.4)
      };
      \addplot[
        black,
        forget plot,
        loosely dotted,
        thin,
      ] coordinates {
        (4.5, -0.15)
        (4.5, 0.4)
      };
      \addplot[
        black,
        forget plot,
        densely dotted,
        thin,
      ] coordinates {
        (5.5, -0.2)
        (5.5, 0.4)
      };
      \addplot[
        color=my_gold,
        error bars/y dir=both,
        error bars/y explicit,
        mark options={fill=my_gold},
        mark=triangle*,
        only marks,
      ] table [
        x expr=\thisrow{cell}-0.21,
        y error minus expr=\thisrow{est}-\thisrow{lo},
        y error plus expr=\thisrow{hi}-\thisrow{est},
        y=est,
      ] {
        cell est lo hi
        1 0.050 0.018 0.092
        2 0.013 -0.031 0.059
        3 -0.037 -0.075 -0.005
        4 0.042 -0.046 0.123
        5 0.059 0.023 0.092
        6 0.052 0.016 0.089
      };
      \addlegendentry{\(\surd\)\textsc{Lerp}}
      \addplot[
        color=my_teal,
        error bars/y dir=both,
        error bars/y explicit,
        mark options={fill=my_teal},
        mark=diamond*,
        only marks,
      ] table [
        x expr=\thisrow{cell}-0.07,
        y error minus expr=\thisrow{est}-\thisrow{lo},
        y error plus expr=\thisrow{hi}-\thisrow{est},
        y=est,
      ] {
        cell est lo hi
        1 0.041 0.021 0.056
        2 0.008 -0.033 0.061
        3 0.006 -0.056 0.079
        4 0.006 -0.083 0.085
        5 0.021 -0.038 0.069
        6 0.043 0.006 0.081
      };
      \addlegendentry{Update merging}
      \addplot[
        color=my_blue,
        error bars/y dir=both,
        error bars/y explicit,
        mark options={fill=my_blue},
        mark=square*,
        only marks,
      ] table[
        x expr=\thisrow{cell}+0.07,
        y error minus expr=\thisrow{est}-\thisrow{lo},
        y error plus expr=\thisrow{hi}-\thisrow{est},
        y=est,
      ] {
        cell est lo hi
        1 0.031 -0.010 0.068
        2 0.009 -0.011 0.027
        3 -0.003 -0.020 0.013
        4 0.063 -0.054 0.171
        5 0.035 -0.021 0.089
        6 0.050 0.015 0.088
      };
      \addlegendentry{\textsc{Ties}}
      \addplot[
        color=my_red,
        error bars/y dir=both,
        error bars/y explicit,
        mark options={fill=my_red},
        mark=pentagon*,
        only marks,
      ] table [
        x expr=\thisrow{cell}+0.21,
        y error minus expr=\thisrow{est}-\thisrow{lo},
        y error plus expr=\thisrow{hi}-\thisrow{est},
        y=est,
      ] {
        cell est lo hi
        1 0.164 -0.008 0.308
        2 0.207 0.128 0.307
        3 0.189 0.107 0.287
        4 0.192 0.107 0.307
        5 0.237 0.123 0.360
        6 0.150 0.087 0.212
      };
      \addlegendentry{Joint}
    \end{axis}
  \end{tikzpicture}

  \pgfplotslegendfromname{clozegridlegend}
  \caption{
    Advantage over the random adapter in cloze retention (mean of ratios).
    Values are shown with \qty{95}{\percent} \acp{ci}, reported in the five cells with \(N \ge 16\) and pooled over all \num{56} collections; \cref{app:clozegrid} gives every cell.
  }\label{fig:clozegrid}
\end{figure*}
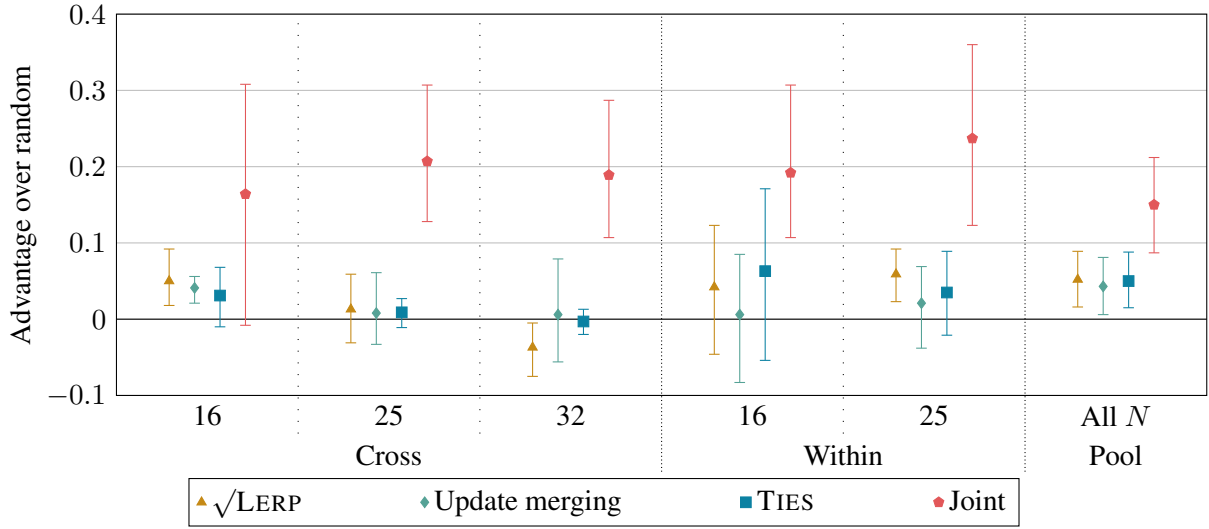

At \num{16} documents or more the merges recover \qtyrange{12}{16}{\percent} of the specialists' cloze advantage over the base model, each \ac{ci} including zero, against \qtyrange{87}{89}{\percent} of their \ac{ob} advantage (\cref{sec:decomposition}).
Paired contrasts place them between the random adapter and joint training (\cref{fig:clozegrid}): pooled over the \num{56} collections, they exceed random on cloze retention by \num{0.043} to \num{0.052}, each \ac{ci} excluding zero, about a third of joint training's \num{0.150} (\ci{0.087}{0.212}), and at \(N \ge 16\) alone \(\num{11} / \num{15}\) contrasts with random include zero, against one of joint training's five.
As aggregate ratios at these sizes the merges keep \numrange{0.80}{0.87} of the specialists' cloze recall and a random adapter \numrange{0.78}{0.83} (close to the base model's \num{0.81} over the same rows), while joint training keeps \numrange{0.955}{1.015} and, since the \ac{ci} on its share reaches \qty{100}{\percent} from below, is not distinguishable from full recovery.

\section{Four silent errors}\label{sec:validity}

Each of these errors produced a clean result we were prepared to report, and none stopped a run or raised an error.

\paragraph{A training-length default.}
In TRL 1.1.0 and 1.2.0 \citep{vonwerra2020trl}, under which our adapters trained, \texttt{SFTConfig.max\_length} defaults to \num{1024} tokens and truncation keeps the start of an over-long string; since our template places the document before the question and answer, in \qty{44}{\percent} of documents no training example kept any token of its answer.
Under that default adapters trained on a document's questions without its text scored \num{0.168} (\ci{0.120}{0.215}) above adapters trained with it \ac{ob}, whereas with a \num{2048}-token window they score \num{0.049} (\ci{0.025}{0.074}) below them, and a domain's change in this gap\---which mirrors what its natural adapters lost\---falls as its truncation rate rises (Spearman \(\rho = \num{-0.943}\), \(p = \num{0.017}\); Pearson \(r = \num{-0.982}\), \(p = \num{0.003}\)) under each of five definitions of a truncated document (\cref{app:truncation}).

\paragraph{A decoding setting in our own code.}
Until we removed it, our generation function applied a repetition penalty of \num{1.1}, which evaluations read from earlier caches kept, so penalized and unpenalized scores sat side by side in our result files; we took the resulting spread for irreproducibility in quantized decoding, whereas under one configuration all \num{150} scores repeat exactly, and penalized denominators had placed update merging's best cell in \cref{sec:merges-ob} above parity, at \num{1.006} against the \num{0.994} of \cref{tab:matchedrank}.

\paragraph{A scaling error in our own merge.}
Our \ac{lerp} implementation multiplied both stored factors by \(1 / N\) before forming their product, so each adapter's own update was scaled to \(1 / N^2\) of its trained magnitude rather than \(1 / N\).
Because the error grows with \(N\), it looked like a mechanism: retention fell smoothly from \num{0.82} of the specialists' score at four documents to \num{0.62} at \num{32} across domains (\cref{app:mergespace}), a curve we read as the cost of merging in factor space until \sqrtlerp{} showed that scale accounts for the whole decline (\cref{sec:mergespace}); the merge of \textsc{Peft} 0.18.1 \citep{mangrulkar2022peft}, which our runs used, forms \sqrtlerp{}'s update exactly, so the error was ours (\cref{app:scale}).

\paragraph{A mismatch in evaluation precision.}
The adapters were trained on a \num{4}-bit base, on which every cloze number in the main text evaluates them unmerged, whereas our first cloze runs had merged each adapter into a bf16 base.
Making the footing uniform lowers the specialists' cloze gain over the base model by \num{0.045} (\ci{0.011}{0.080}) and the synthetic condition's by \num{0.059}, which is then no longer clear of zero; it levels natural training with its style-normalized rewrite and changes whether the \ac{ci} excludes zero for eight of the composition grid's \num{22} contrasts with random (\cref{app:footing}).

\section{Related work}\label{sec:related}

\paragraph{Document adapters and knowledge in weights.}
Parametric \ac{rag} \citep{su2025prag} trains one adapter per passage and, as re-implemented by \citet{tang2025understanding}, sums the updates of those retrieved for a question; re-scored by answer presence, it does not consistently beat retrieval in context, and a generic question-answering adapter trained on \num{400} pooled examples, used as a control, leads the authors to conclude that its gains are mostly document-specific\---a conclusion we reach, with a control drawn from the same population as the specialists (\cref{sec:decomposition}), only on the probe that withholds the document.
\Citet{back2026understanding} find routing among per-document adapters to be the main bottleneck, and \citet{rakotonirina2026where}, asking questions without the document, find that merged per-document adapters degrade as more documents are retrieved while the merges they compare lie within about one point of each other at five documents\---which is consistent with our cloze grid (\cref{sec:cloze-grid}) and with how little the operator matters once scale is right (\cref{sec:mergespace}).
The reading behaviour of \cref{sec:behaviour} fits the account of \citet{zhou2023lima}, in which instruction tuning mostly shapes how capability acquired in pre-training is expressed, and our training conditions (\cref{sec:conditions}) are a document-level case of the finding of \citet{allenzhu2024physics} that whether a fact can be extracted depends on how it is presented in training (further work in \cref{app:related}).

\paragraph{Merging and its scale.}
Task arithmetic \citep{ilharco2023editing} and \ac{ties}-merging \citep{yadav2023ties} act on the full update, whereas a \ac{lora} adapter stores two factors: averaging them instead of their product changes the scale at which each adapter's own update enters as well as adding the cross terms that \citet{prabhakar2025lora} note and \citet{zheng2025decouple} isolate (\cref{sec:mergespace}).

\paragraph{Adapter selection.}
LoraHub \citep{huang2024lorahub} weights adapters on a few target examples, and AdapterFusion \citep{pfeiffer2021adapterfusion} learns attention over several: both presuppose that which adapter is applied matters, a presupposition that at document granularity our cloze probe supports and \ac{ob} scores barely do (\cref{sec:decomposition}).

\section{Conclusion}

On Llama-3.1-8B-Instruct, \ac{ob} scoring mostly credits a reading behaviour every adapter shares: a random adapter recovers about four fifths of the specialists' \ac{ob} advantage over the base model and \qty{2}{\percent} of their cloze advantage, a share not distinguishable from zero.
At \num{16} documents or more, merges that recover \qtyrange{87}{89}{\percent} of the specialists' \ac{ob} advantage recover \qtyrange{12}{16}{\percent} of their cloze advantage, each \ac{ci} including zero, against \qty{94}{\percent} for joint training: so \ac{ob} scores overstate how much of a document's content survives merging or substitution.
Content does reach the weights, whose geometry records what each adapter read; yet in our bank that geometry did not pick out adapters that help with another document's content on either probe, although with thirty queries the cloze \ac{ci} still admits an advantage of about four fifths of what a document's own adapter gains over a random one.

Three recommendations follow:
\begin{enumerate*}[label= (\emph{\alph*})]
  \item score what an adapter holds with a probe that withholds the document, beside the \ac{ob} one;
  \item measure against a random adapter averaged over several draws; and
  \item evaluate specialists, base model, and composed adapters under one precision and decoding configuration, with one set of denominators per table.
\end{enumerate*}
For a collection of up to \num{32} documents known in advance, joint training keeps what merges lose, at the cost of retraining whenever the collection changes.

\section*{Limitations}

Every result except the geometry rests on one base model, Llama-3.1-8B-Instruct, adapted at one rank and under one training schedule on \num{150} encyclopedic articles in six domains, so we cannot say how the shares of \cref{sec:behaviour} move with model size, adapter capacity, or genre; the effect of the training-length default on \ac{ob} scores was likewise measured on this model alone.
That schedule passes over each specialist's eight training strings about a thousand times, with the loss over the whole string, most of whose tokens are the article; such a loss would lead one to expect that the geometry follows the text an adapter read (\cref{sec:cs}), and every adapter learns the shared template from the same repetition.
Composition was tested on collections fixed in advance and no larger than \num{32} documents, where joint training is not yet distinguishable from parity with the specialists (\cref{sec:joint}), so both the size at which a rank-\num{8} joint adapter stops keeping pace and the incremental case, in which composition would have to undercut retraining on each new document, remain untested.

Each document contributes three factual items and one true/false item, enough to estimate means over documents and collections but too few to characterize a single document; this is why the retrieval run cannot resolve content, since over its \num{30} queries even the specialist adapter is not distinguishable from the base model on cloze (\num{+0.085} with \ci{-0.011}{+0.189}; \cref{sec:retrieval}), and an advantage for weight-space selection of up to about four fifths of the own adapter's margin over a random one remains possible.
The two probes also score different items, because cloze shares use the factual items alone, whereas \ac{ob} shares include the true/false item, on which the specialists score below the base model (\cref{app:heldout}); since the items are generated by one model and validated only by string match against the source, a question that admits more than one answer, or a correct fill phrased differently, would go unrecognized, an error we did not measure.

Wikipedia articles have probably been seen in pre-training, a criticism \citet{rakotonirina2026where} raise against evaluating parametric retrieval on such text, and our base model already reaches \num{0.81} of the specialists' cloze recall over the composition grid's rows (\cref{sec:decomposition}); every cloze result is stated against that base model, so what we measure is how adaptation changes the availability of facts the model may already hold, and whether a random adapter recovers as little of the cloze advantage on unseen documents is untested.
All our adapters also share one training format, the instruction template in which \ac{ob} evaluation presents the document (\cref{sec:setup}), so the four fifths that a random adapter recovers measure a behaviour our training instilled in every adapter alike; a bank of adapters trained in different formats might share less of it.
The base model, moreover, is scored \ac{ob} in that raw template rather than in its own chat template, so part of what every adapter adds may be adaptation to the format rather than to reading.

The retrieval bank never holds the query document's own adapter, so that result bears on substituting one document's adapter for another's and says nothing about banks that hold every document's adapter, as parametric \ac{rag} does.
Our account of why the cross terms of \cref{sec:mergespace} cost little here and about ten points across eight vision tasks \citep{zheng2025decouple}\---that updates sharing learned input directions let the cross terms re-add a common component\---is a hypothesis we have not tested; on cloze, moreover, the merges keep so little of the specialists' advantage at \num{16} documents or more that a cost of the cross terms would have little room to~show.

Two checks rest on less evidence than the main results: the footing of the archived \num{4}-bit composition numerators is inferred from the runner's code and supported by \num{64} of the \num{66} evaluations they share with the retrieval run agreeing at six decimals (\cref{app:repeats}), and the check on the chat wrapper rests on \num{35} documents, whose \ac{ci} on the change, \([\num{-0.069}, \num{+0.183}]\), includes a change larger than the gain it checks (\cref{app:template}).
The retrieval contrast is the only comparison whose decision criterion we fixed before computing it, and the other \acp{ci} carry no correction for how many we computed, so a \ac{ci} that only just excludes zero carries little weight by itself.
A cell's \ac{ci} in the composition grid, moreover, is resampled over only five collections, which share documents within and across cells, so it rests on few distinct resamples and will undercover; we therefore take the pooled contrasts as the evidence and the counts of cells whose \acp{ci} exclude zero as~description.

\bibliography{mostly-reading}

\appendix

\section{Setup details}\label{app:setup}

\subsection{Training and software}\label{app:software}

Adapters use rank \num{8}, \(\alpha = 16\) and dropout \num{0.05} on all seven projection matrices, and train for \num{500} steps with AdamW \citep{loshchilov2019decoupled} at a learning rate of \num{2e-4} under a cosine schedule with \num{50} warmup steps and an effective batch size of \num{16}, with the language-modelling loss taken over the whole training string.
Sequences are capped at \num{2048} tokens, which fits every training example, for every adapter we report except the no-context adapters, which trained under the library's default cap of \num{1024} tokens; their training strings carry no article, so that cap does not bear on them, and \cref{sec:validity} describes what the same default did to natural adapters trained before we set the cap ourselves.
Our runs used PyTorch 2.11.0 \citep{ansel2024pytorch}, Transformers 5.5.3 \citep{wolf2020transformers}, \textsc{Peft} 0.18.1 \citep{mangrulkar2022peft}, bitsandbytes 0.49.2~\citep{dettmers2023qlora}, and Accelerate 1.13.0~\citep{gugger2022accelerate}.
Adapters trained under TRL 1.2.0 \citep{vonwerra2020trl}, except the natural adapters trained at the \num{1024}-token window, the Qwen3.5-2B adapters, and the no-context adapters; these trained under TRL 1.1.0, whose defaults and truncation code are the same.
The geometry of \cref{sec:geometry} is replicated with the same hyper-parameters on Llama-3.2-3B-Instruct \citep{metallama2024llama32} and on two models from the Qwen3.5 family \citep{qwenteam2026qwen35}, Qwen3.5-2B \citep{qwen2026qwen352b} and Qwen3.5-9B \citep{qwen2026qwen359b}; every other experiment uses Llama-3.1-8B-Instruct \citep{metallama2024llama31}.
On Llama-3.1-8B-Instruct each adapter has \num{20971520} trainable parameters, about \qty{0.26}{\percent} of the base model's \num{8.0} billion; every run was a single-GPU job on a heterogeneous cluster pool of NVIDIA A100, H100, H200, and B200 GPUs, except the cloze evaluations on the \num{4}-bit footing, which all ran on A100s (\cref{app:repeats}), and we estimate the total budget of all our experiments, superseded and preliminary runs included, at a few hundred GPU hours.

\subsection{Held-out items and the metric}\label{app:heldout}

Every document's held-out set holds three factual questions and one true/false item, and the true/false item does not behave like the others: \ac{ob} the base model answers it correctly for \num{0.987} of documents and the natural adapter for \num{0.933}, so on all four items it narrows the specialists' advantage rather than adding to it, while the swapped-context adapter, scored on its source document with its donor's text in context, answers it correctly for only \num{0.413}, and \ac{cb} every training condition answers it less often than the base model (\numrange{0.520}{0.673} against \num{0.747}).
The cloze probe therefore scores the factual items alone, with the true/false item reported beside them; \ac{ob} scores include it, because the composition and transfer runs keep no per-item \ac{ob} scores, which leaves every \ac{ob} share in this paper on all four items.

Token-level \(F_1\) penalizes answers that are correct but longer than the reference, so a composed adapter that grew more verbose would lose retention without losing knowledge, and we checked for this on a pilot grid of thirty collections, five in each of six cells with \(N \le 8\).
Joint and specialist answers there average \num{10.4} and \num{11.1} words after the metric’s normalization, and retention computed from recall alone, which added tokens cannot lower, exceeds \(F_1\)-based retention by \num{0.002} for joint training and \num{0.005} for \ac{ties}, pooled over the documents of all six cells; since neither the sign nor the size of these differences points to a verbosity artefact, we keep \(F_1\) as the metric.

\subsection{Cloze harness}\label{app:template}

Unlike the \ac{qa} evaluations, the cloze harness wraps each prompt in the model's chat template, which adds the template's default system header and, through the tokenizer, a second beginning-of-sequence token; the wrapper is the same for every arm, the specialist and base-model references included.
Its length caps barely bind: prompts run from \num{60} to \num{108} tokens against a cap of \num{512}, and of the \num{900} cloze generations of the specialists and the base model, \num{8} reach the limit of \num{64} new tokens, all of them the specialists', which can have lowered the specialists' mean recall by at most \(8 / 450 = \num{0.018}\), since only a generation that was cut can lose its fill.
Because the adapters trained without a chat template, we reran \num{35} documents with the training-format prompt instead: under it the specialists return an empty answer on \num{14} of their \num{105} cloze items and the base model on none, which the records keep only as decoded strings, so what the specialists emitted in their place cannot be recovered, and the wrapper does not measurably change the specialists' cloze gain over the base model: chat wrapper minus training format, the gain changes by \num{+0.057} (\ci{-0.069}{+0.183}) with empty answers scored zero and by \num{+0.004} (\ci{-0.147}{+0.151}) on the items both prompts answer, so the subset cannot exclude a change about the size of the gain in either direction.
Those documents are the \num{30} retrieval queries of \cref{sec:retrieval} and five chosen because their specialists recalled none of the factual items on the earlier footing, so their means are not the collection's.

\section{Prompts}\label{app:prompts}

Unicode punctuation is transliterated in the monospace rendering below; the released prompt files carry the original characters.

\subsection{Instruction template}\label{app:prompt-instruction}

The instruction template was fixed and used across all documents, both for training and inference, as
\begin{prompt}
You are a factual question-answering assistant. Answer the question below using only the information provided. Be concise and accurate.

Context: {context}

Question: {question}

Answer:
\end{prompt}

\subsection{Generation of \acl{qa} pairs}\label{app:prompt-qa}

All \ac{qa} pairs, probe items, and document variants were generated with Claude Sonnet 4.6 \citep{anthropic2026introducing} at \num{0.3} temperature with a \num{2048}-token generation limit, using the prompts below. Each generation was validated by string match against the source
document and regenerated on failure.

\paragraph{Base pairs.}
The base \ac{qa} pairs are generated using the system prompt
\begin{prompt}
You are an expert at generating factual question-answer pairs for training language models. Generate high-quality QA pairs strictly based on the provided text.
\end{prompt}
\noindent followed by the user prompt
\begin{prompt}
Read the following text carefully and generate exactly {n_total} question-answer pairs of the following types:
  - {n_factual} single-hop factual questions (direct facts from the text)
  - {n_tf} true/false questions (statement + answer "True" or "False")
  - {n_multihop} multi-hop question (requires connecting two facts from the text)

Rules:
1. Questions must be answerable solely from the provided text.
2. Answers should be concise (1-3 sentences maximum).
3. Avoid trivially easy questions (e.g. "What is the title?").
4. Do NOT ask the same question twice.
5. Vary question phrasing and structure.
6. Return a valid JSON array with this exact schema:
   [
     {{
       "question": "...",
       "answer": "...",
       "type": "factual" | "true_false" | "multi_hop"
     }},
     ...
   ]
   Return ONLY the JSON array, nothing else.

Text:
---
{text}
---
\end{prompt}

\paragraph{Cloze variants.}
The cloze variants are generated using the system prompt
\begin{prompt}
You are a precise QA rewriter. Given a question-answer pair, rewrite it as a fill-in-the-blank cloze sentence.

Rules:
- The cloze sentence must be a complete declarative sentence with exactly one blank (___).
- The blank fills in the key fact stated in the answer.
- The answer to the blank is the shortest phrase that makes the sentence factually correct, matching or closely paraphrasing the original answer.
- Do NOT include any explanation. Output ONLY a JSON object with keys "cloze" and "fill".

Example input:
  question: "What is the boiling point of water at sea level?"
  answer: "The boiling point of water at sea level is 100 degrees Celsius."

Example output:
{"cloze": "The boiling point of water at sea level is ___.", "fill": "100 degrees Celsius"}
\end{prompt}
\noindent followed by the user prompt
\begin{prompt}
question: "{question}"
answer: "{answer}"
\end{prompt}

\subsection{Document variants}\label{app:prompt-variants}

\paragraph{Style-normalized.}
The style-normalized documents are generated using the system prompt
\begin{prompt}
You are an expert technical editor. Your task is to rewrite encyclopedia articles into neutral, consistent expository prose. You must:

1. Preserve ALL factual content: names, dates, quantities, causal relationships, definitions, and examples.
2. Write in plain declarative sentences. Avoid rhetorical questions, vivid metaphors, and stylistic flourishes.
3. Remove Wikipedia-specific patterns: lead-sentence formulas ("X is a Y that..."), "As of [year]", "According to...", citation markers like [1], navbox fragments, and disambiguation notices.
4. Keep approximately the same length as the input (within ±20%).
5. Do not introduce any claims not present in the original.
6. Do not add section headers or bullet points — continuous prose only.
7. Use consistent grammatical register across the entire text (simple present tense for general facts, simple past for historical events).

Return only the rewritten text, nothing else.
\end{prompt}
\noindent followed by the user prompt
\begin{prompt}
Rewrite the following article text in neutral expository prose:

---
{text}
---
\end{prompt}

\paragraph{Synthetic.} The synthetic documents are generated using
the system prompt
\begin{prompt}
You are an information extraction assistant. Given an encyclopedia article, extract its key content and render it as a structured fill-in-the-blank template. You must:

1. Identify the domain (e.g. biology, law, economics).
2. Identify the topic (the main subject of the article in ≤10 words).
3. Write a one-sentence definition of the topic.
4. Extract exactly 6 key facts. Each fact must be a single declarative sentence that states something specific and non-obvious (not just the definition restated). Include names, dates, quantities, causal relationships, or mechanisms where present.
5. Do not invent any information not in the original text.

Return ONLY a JSON object with this exact schema:
{
  "domain": "<string>",
  "topic": "<string>",
  "definition": "<one-sentence string>",
  "key_facts": ["<fact 1>", "<fact 2>", "<fact 3>", "<fact 4>", "<fact 5>", "<fact 6>"]
}
\end{prompt}
\noindent followed by the user prompt
\begin{prompt}
Extract the structured template from the following article:

---
{text}
---
\end{prompt}

\section{Content and geometry}\label{app:content}

\subsection{Training-text conditions}\label{app:conditions}

\Cref{tab:conditionsfull} repeats the scores of \cref{sec:conditions} with every \ac{ci}.
\Ac{ob}, the conditions read different texts: the base model and the natural and no-context adapters read the source document, the two rewrites read their own text, and the swapped adapters read the donor document, scored either with their source's questions, for which no answer is in the context, or with the donor's.

\begin{table*}
  \centering

  \begin{tabular}{l ccc}
    \toprule
    Condition          & \Acl{ob} \(F_1\)            & \Acl{cb} \(F_1\)            & \(\Delta\) cloze recall \\
    \midrule
    Base model         & \estci{0.504}{0.485}{0.522} & \estci{0.320}{0.303}{0.337} & {---} \\
    \midrule
    Natural            & \estci{0.822}{0.803}{0.841} & \estci{0.399}{0.374}{0.424} & \estci{+0.094}{+0.056}{+0.134} \\
    Style-normalized   & \estci{0.708}{0.685}{0.731} & \estci{0.419}{0.394}{0.444} & \estci{+0.097}{+0.054}{+0.141} \\
    Synthetic          & \estci{0.649}{0.621}{0.677} & \estci{0.443}{0.416}{0.470} & \estci{+0.034}{-0.014}{+0.083} \\
    No context         & \estci{0.773}{0.748}{0.796} & \estci{0.426}{0.396}{0.455} & \estci{-0.132}{-0.176}{-0.087} \\
    Swapped, on source & \estci{0.424}{0.394}{0.455} & \estci{0.456}{0.427}{0.484} & \estci{-0.028}{-0.058}{+0.003} \\
    Swapped, on donor  & \estci{0.686}{0.656}{0.715} & \estci{0.428}{0.401}{0.454} & \estci{+0.056}{+0.017}{+0.097} \\
    \bottomrule
  \end{tabular}

  \caption{
    The training-text conditions with every \acs{ci}: \acs{ob} and \acs{cb} token \(F_1\) on all four held-out items and cloze recall on the three factual items as a paired difference from the base model (on the donor document for the donor row).
    Values are computed over all \num{150} documents, with \qty{95}{\percent} percentile bootstrap \acsp{ci} over documents from \num{10000} resamples.
    \Acs{ob} and \acs{cb} use the \num{4}-bit base with greedy decoding, a \num{2048}-token window, no repetition penalty and the training template; cloze uses the \num{4}-bit base with each adapter unmerged and the chat wrapper of \cref{app:template}.
  }\label{tab:conditionsfull}
\end{table*}

\subsection{The context swap on its donor documents}\label{app:swap}

Pairing at the level of the \num{99} distinct donor documents shifts both means slightly from the corpus-wide figures of \cref{tab:conditions}.
On that pairing the swapped adapter reaches \qty{84}{\percent} of the natural adapter's \ac{ob} score on the document it read (\cref{sec:cs}).
On cloze it gains \num{+0.051} (\ci{+0.003}{+0.099}) over the base model on these donors, and the natural adapter leads it by \num{0.039} (\ci{0.001}{0.077}), a \ac{ci} that includes zero once the true/false item is also scored, so we give that margin little weight.

\subsection{Update geometry}\label{app:math}

A \ac{lora} adapter adds to each weight matrix \(\mathbf{W}_0 \in \mathbb{R}^{d_\mathrm{out} \times d_\mathrm{in}}\) the update \(\mathbf{\Delta W} = \mathbf{B} \mathbf{A}\), with \(\mathbf{B} \in \mathbb{R}^{d_\mathrm{out} \times r}\) and \(\mathbf{A} \in \mathbb{R}^{r \times d_\mathrm{in}}\), and our geometry compares updates rather than factors.
For each layer--module pair we compute the truncated \ac{svd} \(\mathbf{\Delta W} = \mathbf{U} \operatorname{diag}(\boldsymbol{\sigma}) \mathbf{V}^\top\) from QR factorizations of \(\mathbf{B}\) and \(\mathbf{A}^\top\), without forming \(\mathbf{\Delta W}\), and normalize the update to unit Frobenius norm by dividing \(\boldsymbol{\sigma}\) by its Euclidean norm, which gives \(\tilde{\boldsymbol{\sigma}}\).

\paragraph{Distance.}
For adapters \(i\) and \(j\) at layer--module pair \(\ell\), the normalized Frobenius distance is
\begin{equation*}
  \begin{split}
    d_\ell(i, j) & = \left\lVert \mathbf{\Delta \tilde{W}}_i^\ell - \mathbf{\Delta \tilde{W}}_j^\ell \right\rVert_F \\
    & = \sqrt{2 - 2 \left\langle \mathbf{\Delta \tilde{W}}_i^\ell, \mathbf{\Delta \tilde{W}}_j^\ell \right\rangle_F}
  \end{split}
\end{equation*}
whose inner product the \ac{svd} factors give as \(\tilde{\boldsymbol{\sigma}}_i^\top (\mathbf{U}_i^\top \mathbf{U}_j \odot \mathbf{V}_i^\top \mathbf{V}_j) \tilde{\boldsymbol{\sigma}}_j\), with \(\odot\) the element-wise product of two \(r \times r\) matrices.
Over a set \(\mathcal{L}\) of layer--module pairs the distance is \(d(i, j) = {(\sum_{\ell \in \mathcal{L}} {d_\ell(i, j)}^2)}^{1/2}\), and since each layer is normalized before aggregation, layers with larger updates do not dominate it.

\paragraph{Subspace overlap.}
The distance compares whole updates; to ask instead whether two adapters occupy the same directions, we compare the column spaces of their left singular vectors,
\begin{equation*}
  o_\ell(i, j) = \frac{\left\lVert \mathbf{U}_i^{\ell\top} \mathbf{U}_j^{\ell} \right\rVert_F}{\sqrt{r}}
\end{equation*}
which equals \num{1} when the two rank-\(r\) subspaces coincide and \num{0} when they are orthogonal, since \(\mathbf{U}\) has orthonormal columns and \(\lVert \mathbf{U}_i^\top \mathbf{U}_j \rVert_F^2 = \sum_k \cos^2\theta_k\) for the principal angles \(\theta_k\) between the two subspaces.
We report the mean of \(o_\ell\) over all \num{224} layer--module combinations.

Two subspaces drawn uniformly at random give an expected overlap of \(\sqrt{r/d}\), which is not a single number here because grouped-query attention gives the projections different output dimensions: \num{0.044} for the query, output, and down projections at \(d = \num{4096}\), \num{0.088} for the key and value projections at \(d = \num{1024}\), and \num{0.024} for the gate and up projections at \(d = \num{14336}\).
Weighting these by the \num{224} layer--module combinations gives a reference of \num{0.051}, which simulation over \num{10000} random frames per shape reproduces to within \qty{1}{\percent}.
The measured \num{0.155} is therefore about three times the random floor.

\paragraph{Update cosine.}
The alignment of the full updates follows from the distances already computed, since for unit-normalized perturbations \(\cos(\mathbf{\Delta W}_i, \mathbf{\Delta W}_j) = 1 - {d_\ell(i,j)}^2/2\) at each layer, and averaging over the \num{224} combinations gives the mean cosine of \cref{sec:geometry}.
The two statistics answer different questions: the subspace overlap asks whether adapters recruit the same directions, and the cosine asks whether the resulting updates point the same way.

\paragraph{A reference for the cosine.}
The cosine also has a reference set by the parametrization of \ac{lora} rather than by chance.
Suppose two rank-\(r\) adapters on a module with input dimension \(d_{\mathrm{in}}\) receive the same gradient \(\mathbf{G}\) with respect to the full weight, and their \texttt{lora\_A} matrices, initialized independently at random, stay near their initial values.
Under plain gradient steps \(\mathbf{B}\) then accumulates in proportion to \(\mathbf{G}\mathbf{A}^\top\), so each update is proportional to \(\mathbf{G}\mathbf{A}^\top\mathbf{A}\), and \(\mathbf{A}^\top\mathbf{A}\) is, up to scale, close to a projection \(\mathbf{P}\) onto a random \(r\)-dimensional subspace of the input space with \(\mathbb{E}[\mathbf{P}] = (r/d_{\mathrm{in}})\mathbf{I}\).
The expected inner product of the two updates is then about \({(r/d_{\mathrm{in}})}^2 \lVert \mathbf{G} \rVert_F^2\) against a squared norm of about \((r/d_{\mathrm{in}}) \lVert \mathbf{G} \rVert_F^2\) for each, a cosine of about \(r/d_{\mathrm{in}}\): \num{0.0020} for the six projections whose input dimension is \num{4096}, \num{0.0006} for the down projection at \num{14336}, and \num{0.0018} averaged over the \num{224} combinations.
A mean above this reference requires the \texttt{lora\_A} matrices of different adapters to have become correlated, that is, to have moved toward shared input directions.

\paragraph{Spread of the cosine.}
Across the \num{11175} pairs the mean cosine of \num{0.0066} has a standard deviation of \num{0.0012}.
Independent random rank-\num{8} updates of the trained shapes give a cosine of zero with a standard deviation of about \num{2e-5}, and the smallest observed pair lies \num{192} of those standard deviations above zero.
We normalize each layer before averaging, so the mean cosine is a statement about the typical layer rather than about the whole update: a global cosine would weight layers by their magnitude, and a few high-norm modules could in principle carry alignment that this average~\mbox{conceals}.

\paragraph{Overlap and semantic similarity.}
Subspace overlap correlates with semantic similarity between the training documents at \(r = \num{0.122}\).
The \num{11175} pairs come from only \num{150} adapters and are therefore not independent, so we assess significance by permuting document labels across adapters rather than across pairs; the observed correlation sits \num{4.4} standard deviations above that null, at \(p < \num{0.001}\) over \num{10000} permutations.
The correlation accounts for roughly \qty{1.5}{\percent} of variance, so adapters trained on closely related documents remain very nearly as orthogonal as adapters trained on unrelated ones.

\subsection{Geometry across architectures}\label{app:geometry}

Distances can be aggregated over every layer--module combination or over a selected subset.
We had fixed a threshold of \(\rho \ge \num{0.250}\) before these runs, intending to select layers clearing it; no single layer does, the strongest reaching \num{0.241}, so we take the highest-correlating layers instead, eight on the base model and between seven and twelve on the others, and call that set late, since on every architecture it falls in the final third of the network.
Llama-3.1-8B's shortfall of \num{0.008} under full aggregation reflects where the threshold was set rather than any measured uncertainty.

Selecting layers by their correlation and then reporting that correlation uses the same data twice.
The held-out \(\rho\) of \cref{sec:geometry} comes from splitting the adapters, choosing layers on one half and evaluating on the other over \num{200} stratified splits; against \num{0.318} in sample its optimism is \num{0.022}, and six of the eight layers are chosen in over \qty{70}{\percent} of splits.
Neighbourhood purity over the selected layers is more sensitive to the selection, falling from \num{0.746} in sample to \num{0.616} held out, so the main text reports purity over all layers, where nothing was selected, with the adapter excluded from its \mbox{own neighbourhood}.

Grouping the \num{150} adapters into six clusters recovers domain labels at an \ac{nmi} of \num{0.511} at full aggregation and \num{0.587} over the selected layers.
The assignments that disagree with the labels are structured rather than scattered: the economics adapters split into a macroeconomic group, holding central banking, fiscal and monetary policy, inflation and the national accounts, and a group holding game theory, market structure and welfare, while the history adapters separate political and military events from social and economic history.

\Cref{tab:geoarch} gives the correlations for all four architectures before and after the truncation fix, together with the clustering metrics and the degree of overlap between the layer sets selected in each case.
The correction barely moves the correlation on any architecture: under full aggregation it drops by at most \num{0.019}, and over the late layers it moves by at most \num{0.021} in either direction, and the layers carrying it are largely the same ones, which is why we treat the geometry result as independent of the training-window error rather than as another of its~casualties.

\begin{table*}
  \centering

  \begin{tabular}{l *{4}{S[table-format=1.3]} *{3}{S[table-format=1.3]} S[table-format=2.0] c}
    \toprule
                                   & \multicolumn{2}{c}{\(\rho\) (all layers)} & \multicolumn{2}{c}{\(\rho\) (late layers)} & {\Acs{nmi}}  & \multicolumn{2}{c}{\Acs{knn} purity} & \multicolumn{2}{c}{Late layers} \\
    \cmidrule(lr){2-3}\cmidrule(lr){4-5}\cmidrule(lr){6-6}\cmidrule(lr){7-8}\cmidrule(lr){9-10}
    \multirow{-2}{*}{Architecture} & {\num{1024}}                              & {\num{2048}}                               & {\num{1024}} & {\num{2048}}                         & {Late} & {Late} & {All} & {\num{2048}} & Kept \\
    \midrule
    Llama-3.1-8B                   & 0.261                                     & 0.242                                      & 0.338        & 0.318                                & 0.587  & 0.746  & 0.708 & 8            & \(4 / 4\) \\
    Llama-3.2-3B                   & 0.434                                     & 0.427                                      & 0.421        & 0.434                                & 0.648  & 0.894  & 0.908 & 7            & \(5 / 8\) \\
    Qwen3.5-2B                     & 0.516                                     & 0.507                                      & 0.518        & 0.540                                & 0.781  & 0.896  & 0.884 & 12           & \(8 / 8\) \\
    Qwen3.5-9B                     & 0.341                                     & 0.333                                      & 0.414        & 0.417                                & 0.609  & 0.876  & 0.823 & 10           & \(7 / 11\) \\
    \bottomrule
  \end{tabular}

  \caption{
    Geometry before and after the truncation fix.
    Spearman correlations are given at both training windows; \acs{nmi} and \acs{knn} purity are reported for the corrected adapters only, \acs{nmi} over the selected late layers and purity both over those layers and over all layers; purity is computed as the fraction of an adapter's six nearest neighbours sharing its domain, for which chance is \num{0.161}; the last two columns give the number of late layers selected after the fix and, of those selected before it, how many are selected again.
  }\label{tab:geoarch}
\end{table*}

\section{Behaviour}\label{app:behaviour}

\subsection{The random-adapter baseline}\label{app:rand}

Pooled over the collections of \num{16} documents or more, the random adapter's \ac{ob} share (\cref{sec:decomposition}) is \num{0.79} (\ci{0.73}{0.85}) across domains and \num{0.87} (\ci{0.81}{0.93}) within one, a figure that includes \(N = 25\), where every random adapter necessarily comes from another domain.

The random baseline averages five independently drawn out-of-collection adapters per collection, and the averaging is needed: individual draws range from \num{0.57} to \num{1.19} of the specialists' score, and at cross-domain \(N = 4\) a single draw returned an adapter at parity with the specialists, against which that cell's contrast with \ac{ties} would no longer have excluded zero.
\Cref{tab:rand} separates the two sources of spread, between the five draws made for a collection and between the collections~\mbox{themselves}.

\begin{table*}
  \centering

  \begin{tabular}{l S[table-format=2.0] S[table-format=1.3] *{4}{S[table-format=1.2]}}
    \toprule
                             &                           &                          & \multicolumn{2}{c}{Standard deviation} & \multicolumn{2}{c}{Draws} \\
    \cmidrule(lr){4-5}\cmidrule(lr){6-7}
    \multirow{-2}{*}{Regime} & {\multirow{-2}{*}{\(N\)}} & {\multirow{-2}{*}{Mean}} & {Within}                               & {Between} & {Min} & {Max} \\
    \midrule
    \multirow{5}{*}{Within}  & 2                         & 0.930                    & 0.09                                   & 0.13      & 0.57  & 1.19 \\
                             & 4                         & 0.917                    & 0.09                                   & 0.04      & 0.70  & 1.08 \\
                             & 8                         & 0.944                    & 0.06                                   & 0.07      & 0.77  & 1.12 \\
                             & 16                        & 0.963                    & 0.04                                   & 0.05      & 0.81  & 1.07 \\
                             & 25                        & 0.968                    & 0.04                                   & 0.02      & 0.84  & 1.04 \\
    \midrule
    \multirow{6}{*}{Cross}   & 2                         & 0.987                    & 0.06                                   & 0.05      & 0.83  & 1.18 \\
                             & 4                         & 0.878                    & 0.06                                   & 0.08      & 0.68  & 1.02 \\
                             & 8                         & 0.963                    & 0.05                                   & 0.05      & 0.80  & 1.10 \\
                             & 16                        & 0.933                    & 0.04                                   & 0.05      & 0.83  & 1.07 \\
                             & 25                        & 0.918                    & 0.05                                   & 0.00      & 0.72  & 1.01 \\
                             & 32                        & 0.941                    & 0.03                                   & 0.04      & 0.83  & 1.06 \\
    \bottomrule
  \end{tabular}

  \caption{
    Mean and spread of the random baseline's \acs{ob} retention.
    \enquote{Within} is the mean standard deviation across the five draws made for a collection, \enquote{between} is the standard deviation of the collection means, and the last two columns the most extreme single draws.
  }\label{tab:rand}
\end{table*}

\subsection{Retrieval by weight-space proximity}\label{app:retrieval}

The criterion and the five-draw analysis of \cref{sec:retrieval} were fixed before the primary contrast was computed: the commit in our repository that declares them fixed is dated 22 September 2026, and the centred arm described below was added the next day, after a two-document pilot on the earlier bf16 footing whose output was discarded without its scores being viewed, and before the full run.

Two further declared contrasts on cloze at top-1 complete those of \cref{sec:retrieval}: weight space against text embedding \num{+0.021} (\ci{-0.036}{+0.077}), and weight space against a random adapter from the query's own domain \num{+0.040} (\ci{-0.020}{+0.098}).

The reference rows fit \cref{sec:decomposition}, a reading we offer as consistent with the data rather than as a prediction we made.
At top-1 the arms that use a foreign adapter lie between \num{-0.027} and \num{+0.016} of the base model on cloze.
A bank without the query's own adapter can help \ac{ob} only through the behaviour every adapter carries, so selection has little to act on there, and on cloze only through content that other documents share, which no method found: weight space and text embedding place their choice in the query's domain for \num{26} and \num{29} of \num{30} queries, yet a random adapter from the query's domain scores no higher on cloze than one from any domain (\num{0.395} against \num{0.415} at top-1), a comparison we did not declare and report only as~\mbox{description}.

Of the \num{24} contrasts, two exclude zero, and they are one contrast under two overlapping scorings: at top-3, where the three highest-ranked adapters are merged with \sqrtlerp{} (\cref{sec:mergespace}), weight space exceeds text embedding on cloze by \num{+0.066} (\ci{+0.009}{+0.123}) on the factual items and by \num{+0.049} (\ci{+0.006}{+0.093}) with the true/false item, because the text-embedding merge falls below both the random merge (\num{0.377} against \num{0.403}) and the base model while the weight-space merge is not distinguishable from random (\num{+0.040} with \ci{-0.017}{+0.099}); we build nothing on it.
The centred variant of weight-space proximity, added on the strength of a diagnostic of hubs that uses no scores and kept outside the decision criterion, is not distinguishable from random (\num{+0.022} with \ci{-0.039}{+0.085}) or from uncentred weight space (\num{+0.003} with \ci{-0.050}{+0.054}).
Chosen for each query with hindsight, the best bank adapter reaches \num{0.953} \ac{ob}, but it is chosen on the items on which it is then scored, so it marks an upper bound rather than a target.
To ask whether a selection rule could approach it, we split each document's held-out items, chose the best adapter on one part and scored that adapter on the other.
Choosing on two factual items and scoring on the third together with the true/false item, the chosen adapter beats a random one by \num{0.026} and \num{0.025} in the two such splits we ran and is also the best on the scoring part for \num{0.27} and \num{0.20} of the \num{30} documents; choosing on one factual item together with the true/false item, which the bank's adapters answer correctly in \num{0.968} of the \num{3600} query--adapter cells, selects on little more than noise, and the chosen adapter is then the best on the scoring part for none of the documents.
Restricted to the factual items, choosing on two and scoring on the third, averaged over the three such splits, the chosen adapter gains \num{0.018} over random selection and is the best for \num{0.18} of the documents, so the ceiling marks no target that a selection rule has been shown to reach.

\begin{table*}
  \centering

  \begin{tabular}{l cc}
    \toprule
    Contrast                                 & {Factual}                      & {With true/false} \\
    \midrule
                                             & \multicolumn{2}{c}{Top 1} \\
    \cmidrule(lr){2-3}
    Weight space \(-\) random (bank)         & \estci{+0.020}{-0.038}{+0.075} & \estci{+0.028}{-0.028}{+0.086} \\
    Weight space \(-\) text embedding        & \estci{+0.021}{-0.036}{+0.077} & \estci{+0.007}{-0.041}{+0.056} \\
    Text embedding \(-\) random (bank)       & \estci{-0.001}{-0.060}{+0.061} & \estci{+0.021}{-0.025}{+0.071} \\
    Weight space \(-\) random (domain)       & \estci{+0.040}{-0.020}{+0.098} & \estci{+0.040}{-0.018}{+0.099} \\
    Centred weight space \(-\) random (bank) & \estci{+0.022}{-0.039}{+0.085} & \estci{+0.022}{-0.032}{+0.079} \\
    Centred \(-\) uncentred weight space     & \estci{+0.003}{-0.050}{+0.054} & \estci{-0.006}{-0.066}{+0.043} \\
    \midrule
                                             & \multicolumn{2}{c}{Top 3} \\
    \cmidrule(lr){2-3}
    Weight space \(-\) random (bank)         & \estci{+0.040}{-0.017}{+0.099} & \estci{+0.025}{-0.016}{+0.070} \\
    Weight space \(-\) text embedding        & \estci{+0.066}{+0.009}{+0.123} & \estci{+0.049}{+0.006}{+0.093} \\
    Text embedding \(-\) random (bank)       & \estci{-0.026}{-0.070}{+0.014} & \estci{-0.024}{-0.064}{+0.013} \\
    Weight space \(-\) random (domain)       & \estci{+0.033}{-0.013}{+0.080} & \estci{+0.026}{-0.006}{+0.061} \\
    Centred weight space \(-\) random (bank) & \estci{+0.020}{-0.028}{+0.071} & \estci{+0.010}{-0.030}{+0.052} \\
    Centred \(-\) uncentred weight space     & \estci{-0.020}{-0.072}{+0.029} & \estci{-0.015}{-0.054}{+0.022} \\
    \bottomrule
  \end{tabular}

  \caption{
    The \num{24} declared cloze contrasts of the retrieval run, paired over the \num{30} query documents, with \qty{95}{\percent} \acsp{ci} from \num{10000} resamples; each query's adapter is selected from a bank of \num{120} that never holds the query's own.
    \enquote{Top~1} applies the selected adapter and \enquote{top~3} merges the three highest-ranked with \(\surd\)\textsc{lerp}; \enquote{factual} scores the three factual items and \enquote{with true/false} adds the fourth; the random arms average five draws.
    The primary contrast is weight space against a random adapter from the bank at top~1 on the factual items, and its decision criterion was fixed before it was computed; the centred contrasts were added afterwards, outside that criterion.
  }\label{tab:retrievalcontrasts}
\end{table*}

\section{Composition}\label{app:composition}

\subsection{Merge operators and scale}\label{app:operators}

Each merge acts independently on each layer--module pair and returns an adapter of the original rank; update merging averages the products \(\mathbf{B}_i \mathbf{A}_i\), whereas \sqrtlerp{} and \ac{ties} act on the stored factors, and the two choices differ in the weight at which each adapter's own update enters the result.

\paragraph{Averaging the updates.}
Let \(\mathbf{\Delta W}_{\Delta} = N^{-1} \sum_i \mathbf{B}_i \mathbf{A}_i\) be the update merge before truncation, and consider one document whose specialist it replaces.
That document's own update survives at \(1 / N\) of its trained magnitude, whereas the other \(N - 1\) updates, being close to orthogonal, sum to a norm about \(\sqrt{N - 1}\) times that of a single update and so contribute about \(\sqrt{N - 1} / N\); the wanted component therefore stands to the rest in a ratio of roughly
\begin{equation*}
  \frac{1 / N}{\sqrt{N - 1} / N} = \frac{1}{\sqrt{N - 1}}
\end{equation*}
which falls from \num{1.00} at \(N = 2\) to \num{0.20} at \(N = 25\).
Orthogonality sets this rate, since updates that opposed one another would grow more slowly in sum, while aligned ones would approach \(N - 1\) times a single update.
On \ac{ob} this dilution costs little, update merging keeping between \num{0.93} and \num{0.99} of the specialists' score over the nine cells with \(N \ge 4\) (\cref{sec:merges-ob}), although on cloze it keeps little of their advantage over the base model (\cref{sec:cloze-grid}).

\paragraph{Returning to rank \texorpdfstring{\(\boldsymbol{r}\)}{r}.}
Stored exactly, the update merge has rank up to \(N r\), since
\begin{equation*}
  \sum_i \mathbf{B}_i \mathbf{A}_i =
  \begin{bmatrix}
    \mathbf{B}_1 & \mathbf{B}_2 & \cdots & \mathbf{B}_N
  \end{bmatrix}
  \begin{bmatrix}
    \mathbf{A}_1 \\
    \mathbf{A}_2 \\
    \vdots \\
    \mathbf{A}_N
  \end{bmatrix}
\end{equation*}
To compare it with the factor merges at equal size we take the truncated \acs{svd} of the averaged update, \(\mathbf{U}_r \operatorname{diag}(\boldsymbol{\sigma}_r) \mathbf{V}_r^\top\), and refactor it as \(\mathbf{B}' = \mathbf{U}_r {\operatorname{diag}(\boldsymbol{\sigma}_r)}^{1/2}\) and \(\mathbf{A}' = {\operatorname{diag}(\boldsymbol{\sigma}_r)}^{1/2} \mathbf{V}_r^\top\), the best rank-\(r\) approximation in Frobenius norm.
The fraction of squared singular values it retains falls from \num{0.72} at \(N = 4\) to \num{0.38} at \(N = 32\), while \ac{ob} retention falls by at most \num{0.034} against the full-rank average (\cref{app:mergespace}).

\paragraph{Averaging the factors.}
Summing each factor over the collection and scaling each sum by \(N^{-1/2}\), as \sqrtlerp{} does, gives
\begin{equation*}
  \begin{gathered}
    \left(\frac{1}{\sqrt{N}} \sum_{i}\mathbf{B}_i\right)\left(\frac{1}{\sqrt{N}} \sum_{j}\mathbf{A}_j\right) = \\
    \frac{1}{N} \sum_{i}\mathbf{B}_i \mathbf{A}_i + \frac{1}{N} \sum_{i \neq j}\mathbf{B}_i \mathbf{A}_j
  \end{gathered}
\end{equation*}
so each adapter's own update enters at \(1 / N\), as in update merging, beside \(N (N - 1)\) cross terms that pair one adapter's output directions with another's input directions, correspond to no trained adapter, and are not constrained by the near-orthogonality of the updates, since that concerns the products \(\mathbf{B}_i \mathbf{A}_i\) and says nothing of \(\mathbf{B}_i \mathbf{A}_j\).
Averaging each factor at \(1 / N\) instead, as our first runs did through the error of \cref{sec:validity}, divides both sums by a further \(N\), so each own update arrives at \(1 / N^2\) of its trained magnitude, a sixty-fourth rather than an eighth at \(N = 8\); comparing \sqrtlerp{} with update merging therefore separates the cost of the cross terms from that of scale (\cref{sec:mergespace}).

\paragraph{\acs{ties}.}
\Ac{ties} \citep{yadav2023ties} applies its three steps to \(\mathbf{A}\) and to \(\mathbf{B}\) separately: it trims each factor to its largest-magnitude entries, elects one sign per parameter by summed magnitude
\begin{equation*}
  \gamma_p = \operatorname{sgn}\left(\sum_{i=1}^{N} \hat{w}_{i,p}\right)
\end{equation*}
where \(\hat{w}_{i,p}\) is the trimmed value of factor entry \(p\) in adapter \(i\), and averages only the entries that agree with the elected sign
\begin{equation*}
  \mathbf{\Delta W}_{\textsc{ties},p} = \frac{1}{\lvert\mathcal{S}_p\rvert} \sum_{i \in \mathcal{S}_p} \hat{w}_{i,p}
\end{equation*}
where \(\mathcal{S}_p = \{ i : \operatorname{sgn}(\hat{w}_{i,p}) = \gamma_p \}\).
Because sign election does not commute with the product, electing over \(\mathbf{A}\) and \(\mathbf{B}\) is not electing over \(\mathbf{B} \mathbf{A}\); a \ac{ties} merge in update space would require materializing each product and refactoring the result, which we do not attempt.

\subsection{Merges at matched rank}\label{app:mergespace}

Against random (\cref{sec:merges-ob}), update merging's contrast at cross-domain \(N = 16\) is the only point estimate below zero, by less than \num{0.001}, and its \ac{ci} excludes zero in six cells of nine.
Against the same average kept at full rank \(N r\), which is equivalent to loading the whole collection, the rank-\num{8} projection of update merging loses between \num{0.002} and \num{0.034} of retention, with six \acp{ci} of nine excluding zero, even though at \(N = 32\) its eight directions carry only \num{0.38} of the summed update's squared singular values (\num{0.72} at \(N = 4\)); whatever the model draws on from these adapters lies in few directions.
A merging coefficient tuned on half of each collection's documents and scored on the other half does worse than \(1 / N\) in four cells, the same in three and better in two, by between \num{-0.061} and \num{+0.019}.
At matched rank \ac{ties} lies above update merging in eight cells of nine (sign test \(p = \num{0.039}\)), and three \acp{ci} exclude zero, two favouring \ac{ties}, by \num{0.020} and \num{0.009}, and one update merging, by \num{0.017}; the largest difference, \num{0.041} at within-domain \(N = 4\), has a \ac{ci} that includes zero.
\Cref{tab:matchedrankfull} gives every method's retention in every cell, the scaling error and joint training included, and \cref{tab:scalecontrasts,tab:baselinecontrasts} the paired contrasts.

\begin{table*}
  \centering

  \begin{tabular}{l S[table-format=2.0] *{8}{S[table-format=1.3]}}
    \toprule
    Regime                  & {\(N\)} & {\makecell{Scaling\\error}} & {\(\surd\)\textsc{Lerp}} & {\makecell{Rank-8\\update}} & {\textsc{Ties}} & {Random} & {Joint} & {\makecell{Full-rank\\update}} & {\makecell{Spectral\\share}} \\
    \midrule
    \multirow{5}{*}{Cross}  & 4       & 0.820                       & 0.996                    & 0.961                       & 0.980           & 0.878    & 1.032   & 0.995                          & 0.719 \\
                            & 8       & 0.713                       & 1.011                    & 0.994                       & 1.004           & 0.963    & 1.027   & 1.013                          & 0.577 \\
                            & 16      & 0.638                       & 0.948                    & 0.932                       & 0.942           & 0.933    & 0.994   & 0.960                          & 0.441 \\
                            & 25      & 0.639                       & 0.978                    & 0.961                       & 0.973           & 0.918    & 1.009   & 0.979                          & 0.390 \\
                            & 32      & 0.622                       & 0.954                    & 0.960                       & 0.968           & 0.941    & 1.023   & 0.974                          & 0.379 \\
    \midrule
    \multirow{4}{*}{Within} & 4       & 0.744                       & 0.982                    & 0.948                       & 0.990           & 0.917    & 1.020   & 0.977                          & 0.716 \\
                            & 8       & 0.677                       & 0.986                    & 0.976                       & 0.985           & 0.944    & 0.988   & 0.979                          & 0.591 \\
                            & 16      & 0.654                       & 0.981                    & 0.989                       & 0.972           & 0.963    & 1.015   & 0.991                          & 0.451 \\
                            & 25      & 0.648                       & 0.996                    & 0.983                       & 0.987           & 0.968    & 1.016   & 0.997                          & 0.404 \\
    \bottomrule
  \end{tabular}

  \caption{
    \Acs{ob} retention at matched rank in the nine cells with \(N \ge 4\).
    Each document's \(F_1\) under the method is divided by that of its own specialist, every column by the same specialists, averaged over documents and then over the cell's sets (five per cell, six at within-domain \(N = 25\)).
    The scaling error is the factor average with both factors scaled by \(1 / N\); the random experiment averages five out-of-collection adapters per set, which at within-domain \(N = 25\) necessarily come from other domains; joint training is a retraining reference; full-rank update keeps the averaged update at rank \(8 N\), a ceiling rather than a merge; the spectral share is the fraction of the averaged update's squared singular values in its top \num{8} directions.
    Joint training at cross-domain \(N = 25\) averages four of the five sets, the fifth having no \acs{ob} evaluation of its joint adapter.
  }\label{tab:matchedrankfull}
\end{table*}

\begin{table*}
  \centering

  \begin{tabular}{l r ccc}
    \toprule
    Regime                  & {\(N\)} & \(\surd\)\textsc{Lerp} \(-\) scaling error & Rank-8 update \(-\) \(\surd\)\textsc{lerp} & Rank-8 \(-\) full-rank update \\
    \midrule
    \multirow{5}{*}{Cross}  & 4       & \estci{+0.176}{+0.126}{+0.240}             & \estci{-0.035}{-0.069}{-0.005}             & \estci{-0.034}{-0.074}{-0.004} \\
                            & 8       & \estci{+0.297}{+0.265}{+0.326}             & \estci{-0.016}{-0.024}{-0.007}             & \estci{-0.019}{-0.030}{-0.010} \\
                            & 16      & \estci{+0.310}{+0.283}{+0.336}             & \estci{-0.015}{-0.046}{+0.008}             & \estci{-0.027}{-0.042}{-0.013} \\
                            & 25      & \estci{+0.338}{+0.322}{+0.358}             & \estci{-0.016}{-0.036}{-0.001}             & \estci{-0.018}{-0.032}{-0.005} \\
                            & 32      & \estci{+0.333}{+0.301}{+0.364}             & \estci{+0.005}{-0.013}{+0.023}             & \estci{-0.014}{-0.020}{-0.010} \\
    \midrule
    \multirow{4}{*}{Within} & 4       & \estci{+0.238}{+0.174}{+0.307}             & \estci{-0.033}{-0.078}{+0.011}             & \estci{-0.029}{-0.079}{+0.011} \\
                            & 8       & \estci{+0.309}{+0.252}{+0.360}             & \estci{-0.009}{-0.023}{+0.004}             & \estci{-0.002}{-0.022}{+0.017} \\
                            & 16      & \estci{+0.327}{+0.289}{+0.370}             & \estci{+0.008}{-0.010}{+0.032}             & \estci{-0.002}{-0.008}{+0.006} \\
                            & 25      & \estci{+0.348}{+0.322}{+0.375}             & \estci{-0.013}{-0.018}{-0.009}             & \estci{-0.015}{-0.021}{-0.007} \\
    \bottomrule
  \end{tabular}

  \caption{
    Paired \acs{ob} contrasts that isolate scale, cross terms, and truncation.
    All values are shown with \qty{95}{\percent} \acsp{ci}, computed from \num{10000} resamples over sets.
    \(\surd\)\textsc{Lerp} minus the scaling error isolates scale, rank-\num{8} update minus \(\surd\)\textsc{lerp} isolates cross terms at equal own-term weight, and rank-\num{8} update minus full-rank update isolates the cost of truncation.
    Each contrast is computed from unrounded retentions, so it can differ in its last digit from the difference of the displayed values.
  }\label{tab:scalecontrasts}
\end{table*}

\begin{table*}
  \centering

  \begin{tabular}{l r ccc}
    \toprule
    Regime                  & {\(N\)} & \(\surd\)\textsc{Lerp} \(-\) random & Rank-8 update \(-\) random     & \textsc{Ties} \(-\) rank-8 update \\
    \midrule
    \multirow{5}{*}{Cross}  & 4       & \estci{+0.117}{+0.061}{+0.205}      & \estci{+0.082}{+0.029}{+0.139} & \estci{+0.020}{+0.003}{+0.037} \\
                            & 8       & \estci{+0.047}{+0.018}{+0.072}      & \estci{+0.031}{+0.009}{+0.053} & \estci{+0.009}{+0.001}{+0.016} \\
                            & 16      & \estci{+0.015}{-0.002}{+0.045}      & \estci{-0.000}{-0.008}{+0.011} & \estci{+0.010}{-0.007}{+0.021} \\
                            & 25      & \estci{+0.060}{+0.033}{+0.083}      & \estci{+0.043}{+0.017}{+0.067} & \estci{+0.011}{-0.002}{+0.027} \\
                            & 32      & \estci{+0.014}{-0.000}{+0.029}      & \estci{+0.019}{+0.014}{+0.025} & \estci{+0.008}{-0.006}{+0.030} \\
    \midrule
    \multirow{4}{*}{Within} & 4       & \estci{+0.065}{+0.015}{+0.136}      & \estci{+0.031}{-0.015}{+0.077} & \estci{+0.041}{-0.006}{+0.121} \\
                            & 8       & \estci{+0.041}{+0.024}{+0.058}      & \estci{+0.032}{+0.027}{+0.036} & \estci{+0.009}{-0.002}{+0.020} \\
                            & 16      & \estci{+0.018}{-0.007}{+0.052}      & \estci{+0.026}{-0.002}{+0.058} & \estci{-0.017}{-0.032}{-0.003} \\
                            & 25      & \estci{+0.028}{+0.021}{+0.037}      & \estci{+0.015}{+0.008}{+0.021} & \estci{+0.004}{-0.007}{+0.016} \\
    \bottomrule
  \end{tabular}

  \caption{
    Paired \acs{ob} contrasts of the correctly scaled merges with the random adapter and of \textsc{Ties} with the rank-\num{8} update.
    Values are shown with \qty{95}{\percent} \acsp{ci} from \num{10000} resamples over sets.
    Each contrast is computed from unrounded retentions, so it can differ in its last digit from the difference of the displayed values.
  }\label{tab:baselinecontrasts}
\end{table*}

\paragraph{Prior measurements of the cross terms.}
\Citet{tang2025lora} measure the gap between merging the factors and merging the updates of rank-\num{16} adapters on eight vision tasks and attribute it to the cross terms; under averaging, though, their factor merge also enters each adapter's own update at \(1 / N^2\) rather than \(1 / N\), and their control, which adds noise matched to the cross terms' mean and standard deviation to update merging, keeps update merging's own-term scale, so the gap they report joins scale to cross terms.
\Citet{zheng2025decouple} write their separate merge with one coefficient on the own terms and the cross terms alike, which as written isolates the cross terms as \sqrtlerp{} does, and find that they cost about ten points across eight vision tasks, against at most \num{0.035} of \ac{ob} retention across our documents; KnOTS \citep{stoica2025model} argues that cross terms can cost performance and merges only full updates, but its merged update is not limited in rank and so offers no matched-budget comparison.
Our updates share input directions learned in training (\cref{sec:geometry}), so a cross term \(\mathbf{B}_i \mathbf{A}_j\) may partly re-add what every update carries, which would make the cost of cross terms depend on how alike the merged adapters are; we offer this as a hypothesis we have not tested.

\subsection{Joint training and the full grid}\label{app:grid}

Per cell, joint training's lead over the merges (\cref{sec:joint}) is less uniform than pooled: \ac{ob}, its \ac{ci} includes zero against \sqrtlerp{} and \ac{ties} at cross-domain \(N = 25\), against \sqrtlerp{} at within-domain \(N = 25\) and against update merging at within-domain \(N = 16\) (\cref{tab:composition}), as it does in the within-domain cells at \(N = 4\) and \(N = 8\), and at within-domain \(N = 2\) joint training sits below every merge on both probes.
At cross-domain \(N = 32\), where \num{49} of the \num{150} documents appear in more than one collection, joint training is not distinguishable from parity with the specialists, while its lead over random excludes zero under every resampling scheme we tried, over collections and over~\mbox{documents}.

\begin{table*}
  \centering

  \begin{tabular}{l r ccc}
    \toprule
                            & {\(N\)} & Joint \(-\) \(\surd\)\textsc{lerp} & Joint \(-\) rank-8 update      & Joint \(-\) \textsc{ties} \\
    \midrule
    \multirow{3}{*}{Cross}  & 16      & \estci{+0.046}{+0.021}{+0.071}     & \estci{+0.062}{+0.023}{+0.087} & \estci{+0.052}{+0.007}{+0.084} \\
                            & 25      & \estci{+0.020}{-0.006}{+0.046}     & \estci{+0.043}{+0.011}{+0.074} & \estci{+0.029}{-0.007}{+0.065} \\
                            & 32      & \estci{+0.068}{+0.049}{+0.088}     & \estci{+0.063}{+0.051}{+0.072} & \estci{+0.055}{+0.032}{+0.075} \\
    \midrule
    \multirow{2}{*}{Within} & 16      & \estci{+0.034}{+0.000}{+0.084}     & \estci{+0.026}{-0.001}{+0.056} & \estci{+0.043}{+0.010}{+0.080} \\
                            & 25      & \estci{+0.020}{-0.005}{+0.041}     & \estci{+0.033}{+0.009}{+0.051} & \estci{+0.029}{+0.011}{+0.043} \\
    \midrule
    Pooled                  & {---}   & \estci{+0.038}{+0.023}{+0.053}     & \estci{+0.045}{+0.032}{+0.058} & \estci{+0.042}{+0.027}{+0.056} \\
    \bottomrule
  \end{tabular}

  \caption{
    Paired contrasts in retention of joint training against each rank-\num{8} merge at \(N \ge 16\), \acs{ob}.
    Values are computed at set level, with \qty{95}{\percent} \acsp{ci} from \num{10000} resamples.
    Joint training at cross-domain \(N = 25\) averages four of the five sets, the fifth having no \acs{ob} evaluation of its joint adapter.
    Each contrast is computed from unrounded retentions, so it can differ in its last digit from the difference of the displayed values.
  }\label{tab:composition}
\end{table*}

\paragraph{Training budget.}
Each joint adapter trains for the same \num{500} steps as each specialist it replaces, so at \(N\) documents it receives a factor of \(N\) less gradient per document.
Repeating \(N = 16\) with the budget scaled linearly to \num{8000} steps, so that each document receives the gradient it received as a specialist, retention at the matched budget minus that at the scaled one is \num{+0.017} (\ci{-0.010}{+0.049}) within domains (\num{1.015} against \num{0.998}) and \num{+0.006} (\ci{-0.011}{+0.018}) across them (\num{0.994} against \num{0.988}); neither \ac{ci} excludes zero, which places joint training's flat retention on capacity rather than on undertraining.

\paragraph{The economics deficit.}
Economics is the one domain whose \(N = 25\) joint adapter falls below parity, at \num{0.940}, and its deficit is concentrated rather than diffuse: two documents, price elasticity of demand (\num{0.562}) and fiscal policy (\num{0.603}), lose most, and without them the domain's mean rises to \num{0.971}, while their specialist scores repeat exactly under the corrected configuration, so they are not artefacts of the denominator; we report the deficit without a~\mbox{mechanism}.

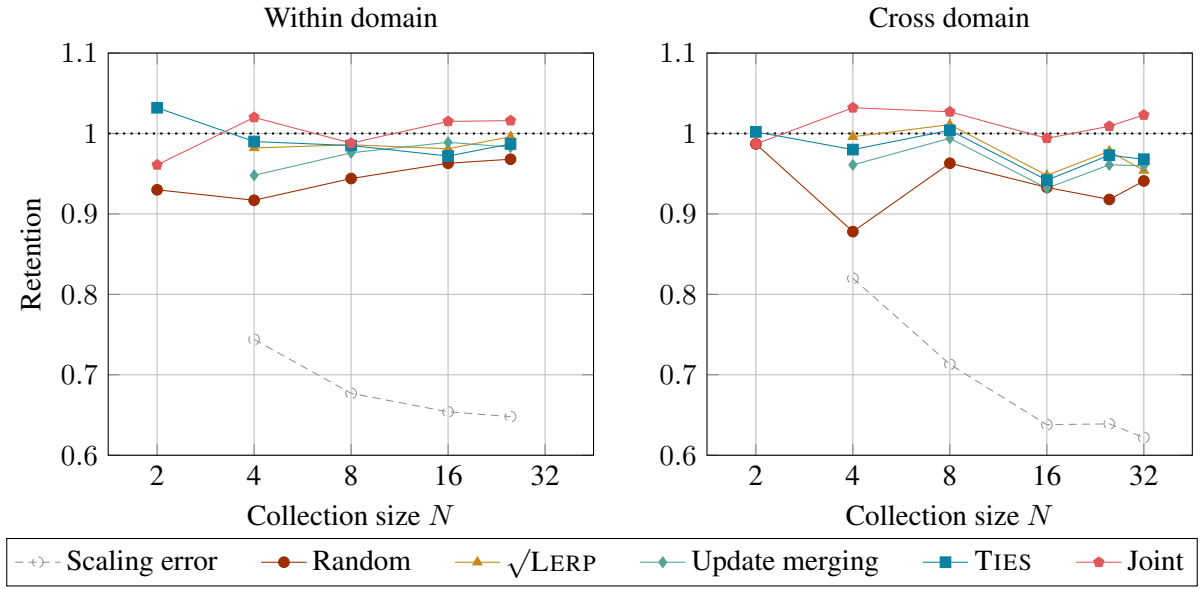
\begin{figure*}[t]
  \centering

  \begin{tikzpicture}
    \begin{groupplot}[
        width=0.5\textwidth,
        grid=major,
        group style={
          group size=2 by 1,
          horizontal sep=15mm,
        },
        legend style={
          /tikz/every even column/.style={column sep=5mm},
          legend columns=-1,
        },
        log basis x=2,
        xlabel={Collection size \(N\)},
        xmax=45.25,
        xmin=1.41,
        xmode=log,
        xtick={
          2,
          4,
          8,
          16,
          32
        },
        xticklabels={
          2,
          4,
          8,
          16,
          32
        },
        ymax=1.1,
        ymin=0.6,
        ytick={
          0.6,
          0.7,
          0.8,
          0.9,
          1.0,
          1.1
        },
      ]

      \nextgroupplot[
        title={Within domain},
        ylabel={Retention},
      ]
      \addplot[
        color=black,
        dotted,
        mark=none,
        thick,
      ] coordinates {
        (1.41, 1)
        (45.25, 1)
      };
      \addplot[
        color=my_gray,
        densely dashed,
        mark=o,
      ] coordinates {
        (4,0.744)
        (8,0.677)
        (16,0.654)
        (25,0.648)
      };
      \addplot[
        color=my_darkred,
        mark=*,
      ] coordinates {
        (2,0.930)
        (4,0.917)
        (8,0.944)
        (16,0.963)
        (25,0.968)
      };
      \addplot[
        color=my_gold,
        mark=triangle*,
      ] coordinates {
        (4,0.982)
        (8,0.986)
        (16,0.981)
        (25,0.996)
      };
      \addplot[
        color=my_teal,
        mark=diamond*,
      ] coordinates {
        (4,0.948)
        (8,0.976)
        (16,0.989)
        (25,0.983)
      };
      \addplot[
        color=my_blue,
        mark=square*,
      ] coordinates {
        (2,1.032)
        (4,0.990)
        (8,0.985)
        (16,0.972)
        (25,0.987)
      };
      \addplot[
        color=my_red,
        mark=pentagon*,
      ] coordinates {
        (2,0.961)
        (4,1.020)
        (8,0.988)
        (16,1.015)
        (25,1.016)
      };

      \nextgroupplot[
        legend to name={retentionlegend},
        title={Cross domain},
      ]
      \addplot[
        color=black,
        dotted,
        forget plot,
        mark=none,
        thick,
      ] coordinates {
        (1.41,1)
        (45.25,1)
      };
      \addplot[
        color=my_gray,
        densely dashed,
        mark=o,
      ] coordinates {
        (4,0.820)
        (8,0.713)
        (16,0.638)
        (25,0.639)
        (32,0.622)
      };
      \addlegendentry{Scaling error}
      \addplot[
        color=my_darkred,
        mark=*,
      ] coordinates {
        (2,0.987)
        (4,0.878)
        (8,0.963)
        (16,0.933)
        (25,0.918)
        (32,0.941)
      };
      \addlegendentry{Random}
      \addplot[
        color=my_gold,
        mark=triangle*,
      ] coordinates {
        (4,0.996)
        (8,1.011)
        (16,0.948)
        (25,0.978)
        (32,0.954)
      };
      \addlegendentry{\(\surd\)\textsc{Lerp}}
      \addplot[
        color=my_teal,
        mark=diamond*,
      ] coordinates {
        (4,0.961)
        (8,0.994)
        (16,0.932)
        (25,0.961)
        (32,0.960)
      };
      \addlegendentry{Update merging}
      \addplot[
        color=my_blue,
        mark=square*,
      ] coordinates {
        (2,1.002)
        (4,0.980)
        (8,1.004)
        (16,0.942)
        (25,0.973)
        (32,0.968)
      };
      \addlegendentry{\textsc{Ties}}
      \addplot[
        color=my_red,
        mark=pentagon*,
      ] coordinates {
        (2,0.987)
        (4,1.032)
        (8,1.027)
        (16,0.994)
        (25,1.009)
        (32,1.023)
      };
      \addlegendentry{Joint}
    \end{groupplot}
  \end{tikzpicture}

  \pgfplotslegendfromname{retentionlegend}
  \caption{
    \Ac{ob} retention against collection size.
    \Acsp{ci} are omitted for legibility, so the paired contrasts of \cref{app:mergespace,app:grid}, rather than the figure, carry the claims about significance; the \num{1.0} marks parity with the per-document specialist each method replaces.
    Scaling error is the factor average at \(1 / N^2\); joint training at cross-domain \(N = 25\) averages four of the five sets, the fifth having no \acs{ob} evaluation of its joint adapter.
  }\label{fig:retention}
\end{figure*}

\paragraph{Retention by collection size.}
\Cref{fig:retention} plots retention against collection size for every method; \sqrtlerp{}, update merging and the scaling error are drawn from \(N = 4\), the range of the matched-rank comparison, since they were not run \ac{ob} at \(N = 2\), and the within-domain regime stops at \(N = 25\) because each domain holds exactly that many~\mbox{documents}.

\subsection{Cloze composition grid}\label{app:clozegrid}

At \(N \ge 16\) the merges' cloze contrasts with random run from \num{-0.037} to \num{+0.063}, with \sqrtlerp{} excluding zero in three cells, two above the random adapter and one (cross-domain \(N = 32\)) below it, update merging in one and \ac{ties} in none, while joint training leads random by \num{0.164} to \num{0.237}, four \acp{ci} of five excluding zero.
At within-domain \(N = 2\) joint training sits below the random adapter on cloze (\num{-0.242} with \ci{-0.450}{-0.033}), as it sits below every merge there on both probes (\cref{app:grid}); we report this without an explanation.
\Cref{tab:clozegrid} gives both aggregations of retention for every cell and method, and \cref{tab:clozecontrasts} each method's advantage over random; \sqrtlerp{} and update merging were run at \(N = 2\) on cloze only.
The seven documents whose specialist recalls none of the factual items, and which the mean of ratios therefore leaves out, are adenosine triphosphate, algorithm, gradient descent, sorting algorithm, microeconomics, the Industrial Revolution and judicial review.
The random, \ac{ties} and joint numerators come from an earlier run whose result files do not record the base model's precision; its code loads the base in \num{4}-bit, and \num{64} of the \num{66} evaluations it shares with the retrieval run agree to six decimals, so we treat it as on the same footing.

\begin{table*}
  \centering

  \begin{tabular}{l S[table-format=2.0] *{10}{S[table-format=1.3]}}
    \toprule
                             &                           & \multicolumn{2}{c}{Random} & \multicolumn{2}{c}{\(\surd\)\textsc{Lerp}} & \multicolumn{2}{c}{Rank-8 update} & \multicolumn{2}{c}{\textsc{Ties}} & \multicolumn{2}{c}{Joint} \\
    \cmidrule(lr){3-4}\cmidrule(lr){5-6}\cmidrule(lr){7-8}\cmidrule(lr){9-10}\cmidrule(lr){11-12}
    \multirow{-2}{*}{Regime} & {\multirow{-2}{*}{\(N\)}} & {MR}                       & {AR}                                       & {MR}                              & {AR}                              & {MR}  & {AR}  & {MR}  & {AR}  & {MR}  & {AR} \\
    \midrule
    \multirow{6}{*}{Cross}   & 2                         & 0.916                      & 0.876                                      & 1.014                             & 1.005                             & 1.086 & 1.056 & 1.089 & 1.065 & 1.103 & 1.066 \\
                             & 4                         & 1.017                      & 0.901                                      & 1.154                             & 1.044                             & 1.104 & 0.987 & 1.142 & 1.044 & 1.197 & 1.139 \\
                             & 8                         & 0.966                      & 0.873                                      & 0.975                             & 0.919                             & 0.930 & 0.847 & 0.916 & 0.853 & 1.195 & 1.085 \\
                             & 16                        & 0.893                      & 0.811                                      & 0.943                             & 0.849                             & 0.934 & 0.853 & 0.924 & 0.842 & 1.057 & 0.966 \\
                             & 25                        & 0.896                      & 0.818                                      & 0.909                             & 0.826                             & 0.904 & 0.826 & 0.905 & 0.827 & 1.103 & 0.993 \\
                             & 32                        & 0.923                      & 0.817                                      & 0.886                             & 0.796                             & 0.929 & 0.831 & 0.920 & 0.819 & 1.112 & 1.000 \\
    \midrule
    \multirow{5}{*}{Within}  & 2                         & 1.057                      & 1.012                                      & 1.148                             & 1.118                             & 1.214 & 1.133 & 1.164 & 1.089 & 0.815 & 0.818 \\
                             & 4                         & 0.886                      & 0.873                                      & 0.975                             & 0.956                             & 0.880 & 0.877 & 0.977 & 0.958 & 1.040 & 0.994 \\
                             & 8                         & 0.977                      & 0.868                                      & 0.997                             & 0.943                             & 1.000 & 0.925 & 0.949 & 0.903 & 1.115 & 1.017 \\
                             & 16                        & 0.860                      & 0.782                                      & 0.901                             & 0.826                             & 0.866 & 0.801 & 0.922 & 0.847 & 1.052 & 0.955 \\
                             & 25                        & 0.895                      & 0.827                                      & 0.953                             & 0.872                             & 0.915 & 0.849 & 0.930 & 0.864 & 1.131 & 1.015 \\
    \midrule
    Pooled                   & {---}                     & 0.934                      & 0.827                                      & 0.986                             & 0.857                             & 0.977 & 0.851 & 0.984 & 0.858 & 1.085 & 1.000 \\
    \bottomrule
  \end{tabular}

  \caption{
    Cloze retention relative to each document's own specialist, on the \num{4}-bit footing.
    \enquote{MR} is the mean over documents of the ratio of recalls, without the seven documents whose specialist recalls none of the factual items; \enquote{AR} is the ratio of mean recalls, with them; the pooled AR covers \num{735} rows.
    The random, \textsc{ties}, and joint values come from an earlier run, whose footing is described in \cref{app:clozegrid}.
  }\label{tab:clozegrid}
\end{table*}

\begin{sidewaystable*}[p]
  \centering

  \begin{tabular}{l r cccc}
    \toprule
    Regime                               & {\(N\)}                        & \(\surd\)\textsc{Lerp} \(-\) random & Rank-8 update \(-\) random     & \textsc{Ties} \(-\) random     & Joint \(-\) random \\
    \midrule
    \multirow{6}{*}{Cross}               & 2                              & \estci{+0.099}{-0.022}{+0.219}      & \estci{+0.170}{+0.102}{+0.246} & \estci{+0.174}{+0.076}{+0.273} & \estci{+0.187}{-0.020}{+0.409} \\
                                         & 4                              & \estci{+0.137}{+0.000}{+0.302}      & \estci{+0.087}{-0.050}{+0.277} & \estci{+0.125}{-0.032}{+0.324} & \estci{+0.180}{-0.064}{+0.412} \\
                                         & 8                              & \estci{+0.009}{-0.107}{+0.107}      & \estci{-0.035}{-0.134}{+0.053} & \estci{-0.050}{-0.108}{+0.004} & \estci{+0.229}{+0.060}{+0.372} \\
                                         & 16                             & \estci{+0.050}{+0.018}{+0.092}      & \estci{+0.041}{+0.021}{+0.056} & \estci{+0.031}{-0.010}{+0.068} & \estci{+0.164}{-0.008}{+0.308} \\
                                         & 25                             & \estci{+0.013}{-0.031}{+0.059}      & \estci{+0.008}{-0.033}{+0.061} & \estci{+0.009}{-0.011}{+0.027} & \estci{+0.207}{+0.128}{+0.307} \\
                                         & 32                             & \estci{-0.037}{-0.075}{-0.005}      & \estci{+0.006}{-0.056}{+0.079} & \estci{-0.003}{-0.020}{+0.013} & \estci{+0.189}{+0.107}{+0.287} \\
    \midrule
    \multirow{5}{*}{Within}              & 2                              & \estci{+0.091}{-0.122}{+0.271}      & \estci{+0.158}{-0.067}{+0.382} & \estci{+0.108}{-0.024}{+0.240} & \estci{-0.242}{-0.450}{-0.033} \\
                                         & 4                              & \estci{+0.089}{-0.058}{+0.276}      & \estci{-0.006}{-0.117}{+0.104} & \estci{+0.091}{-0.049}{+0.280} & \estci{+0.154}{-0.106}{+0.448} \\
                                         & 8                              & \estci{+0.020}{-0.132}{+0.133}      & \estci{+0.022}{-0.118}{+0.119} & \estci{-0.029}{-0.175}{+0.092} & \estci{+0.138}{-0.018}{+0.270} \\
                                         & 16                             & \estci{+0.042}{-0.046}{+0.123}      & \estci{+0.006}{-0.083}{+0.085} & \estci{+0.063}{-0.054}{+0.171} & \estci{+0.192}{+0.107}{+0.307} \\
                                         & 25                             & \estci{+0.059}{+0.023}{+0.092}      & \estci{+0.021}{-0.038}{+0.069} & \estci{+0.035}{-0.021}{+0.089} & \estci{+0.237}{+0.123}{+0.360} \\
    \midrule
    \multicolumn{2}{l}{Pooled, \num{56}} & \estci{+0.052}{+0.016}{+0.089} & \estci{+0.043}{+0.006}{+0.081}      & \estci{+0.050}{+0.015}{+0.088} & \estci{+0.150}{+0.087}{+0.212} \\
    \bottomrule
  \end{tabular}

  \caption{
    Each method's advantage over the random adapter in cloze retention.
    All values are mean of ratios, paired over collections, with \qty{95}{\percent} \acsp{ci} from \num{10000} resamples.
    Each contrast is computed from unrounded retentions, so it can differ in its last digit from the difference of the displayed values.
  }\label{tab:clozecontrasts}
\end{sidewaystable*}

\subsection{Pairwise interference}\label{app:interference}

Merging pairs of adapters with \ac{ties}, over \num{200} pairs stratified into four equal-quantile bins of semantic distance between the two training documents, changes accuracy by \num{-0.011}, \num{-0.007}, \num{+0.001} and \num{-0.016} from the closest bin to the most distant, whose mean distances are \num{0.736}, \num{0.861}, \num{0.917} and \num{0.977}.
Neither distance anticipates the change: its correlation with semantic distance and with weight-space distance stays within \(|r| \le \num{0.016}\), and between weight-space distance and the change Spearman \(\rho = \num{0.000}\) with \(p = \num{0.999}\) under a permutation of adapter labels, which respects the dependence between pairs that share an adapter.
Within-domain pairs lose more than cross-domain ones (\num{-0.026} against \num{-0.005}), the opposite of what a distance that anticipated interference would give.
We report no factor-averaging merge here, since the pairwise \ac{lerp} merges of these runs carried the scaling error described in \cref{sec:validity}.

\section{Silent errors}\label{app:measurement}

\subsection{Truncation by domain}\label{app:truncation}

Under the training-length default of \cref{sec:validity}, the share of documents in which none of the eight training examples kept any token of its answer, counted with the training tokenizer, runs from \qty{4}{\percent} in law to \qty{80}{\percent} in biology, \qty{44}{\percent} overall (\enquote{all answers} row of \cref{tab:truncdefs}).
\Cref{tab:truncation} gives each domain's gap between no-context and natural adapters at both training windows: the sign changes in four of the six domains, since law is negative at both windows and medicine sits at \num{+0.001} (\ci{-0.064}{+0.063}) at \num{2048}, and of the five domains negative at \num{2048} only economics and law exclude zero.
The correlation that \cref{sec:validity} reports holds under each of the five definitions of \cref{tab:truncdefs}, with Spearman \(\rho\) between \num{-0.943} and \num{-0.986} and Pearson \(r\) between \num{-0.960} and \num{-0.982}.
Evaluation had a window of its own, which in some of our earlier runs was \num{512} tokens and cut every \ac{ob} document to about \num{330} of its \numrange{871}{1257} tokens; at the \num{2048} tokens used throughout, no prompt is cut.

\begin{table*}
  \centering

  \begin{tabular}{l *{6}{S[table-format=2.0]} S[table-format=2.1] S[table-format=+1.3] S[table-format=1.3]@{}l S[table-format=+1.3] S[table-format=1.3]@{}l}
    \toprule
                                 & \multicolumn{7}{c}{Truncated documents (\%)} & \multicolumn{3}{c}{Spearman} & \multicolumn{3}{c}{Pearson} \\
    \cmidrule(lr){2-8}\cmidrule(lr){9-11}\cmidrule(lr){12-14}
    \multirow{-2}{*}{Definition} & {L}                                          & {E}                          & {CS} & {H} & {M} & {B} & {All} & {\(\rho\)} & \multicolumn{2}{c}{\(p\)} & {\(r\)}   & \multicolumn{2}{c}{\(p\)} \\
    \midrule
    First answer                 & 8                                            & 24                           & 44   & 56  & 64  & 80  & 46.0  & -0.943     & 0.017                     & \(^{*}\)  & -0.976 & 0.003 & \(^{**}\) \\
    Any string                   & 40                                           & 48                           & 68   & 84  & 88  & 96  & 70.7  & -0.943     & 0.017                     & \(^{*}\)  & -0.960 & 0.003 & \(^{**}\) \\
    Any answer                   & 12                                           & 28                           & 48   & 72  & 72  & 84  & 52.7  & -0.986     & 0.006                     & \(^{**}\) & -0.963 & 0.003 & \(^{**}\) \\
    All answers                  & 4                                            & 24                           & 40   & 56  & 60  & 80  & 44.0  & -0.943     & 0.017                     & \(^{*}\)  & -0.982 & 0.003 & \(^{**}\) \\
    Mean share                   & 6                                            & 24                           & 44   & 58  & 64  & 82  & 46.6  & -0.943     & 0.017                     & \(^{*}\)  & -0.980 & 0.003 & \(^{**}\) \\
    \bottomrule
  \end{tabular}

  \caption[Truncation under five definitions]{
    Truncation at the \num{1024}-token default under five definitions, counted with the training tokenizer, and each definition's correlation across the six domains with the change in the gap (\cref{tab:truncation}).
    A document counts as truncated when:
    \begin{enumerate*}[label= (\emph{\alph*})]
      \item the answer of its first training example lies wholly beyond the window;
      \item any of its training strings exceeds the window;
      \item at least one of its answers lies wholly beyond it;
      \item all eight do, the definition of \cref{sec:validity}; or
      \item is the mean share of the eight answers that do.
    \end{enumerate*}
    \enquote{L} stands for law, \enquote{E} for economics, \enquote{CS} for computer science, \enquote{H} for history, \enquote{M} for medicine, and \enquote{B} for biology.
    The \mbox{\(p\)-values} are computed over the \num{720} orderings; \(^{*} p < \num{0.05}\), \mbox{\(^{**} p < \num{0.01}\), \(^{***} p < \num{0.001}\)}.
  }\label{tab:truncdefs}
\end{table*}

\begin{table*}
  \centering

  \begin{tabular}{l ccc}
    \toprule
                             & \multicolumn{2}{c}{Gap}        & \\
    \cmidrule(lr){2-3}
    \multirow{-2}{*}{Domain} & at \num{1024}                  & at \num{2048}                  & \multirow{-2}{*}{Change} \\
    \midrule
    Law                      & \estci{-0.017}{-0.105}{+0.073} & \estci{-0.078}{-0.137}{-0.017} & \estci{-0.061}{-0.122}{-0.005} \\
    Economics                & \estci{+0.042}{-0.054}{+0.144} & \estci{-0.071}{-0.130}{-0.013} & \estci{-0.112}{-0.201}{-0.031} \\
    C.\ science              & \estci{+0.148}{+0.010}{+0.275} & \estci{-0.062}{-0.134}{+0.002} & \estci{-0.210}{-0.302}{-0.121} \\
    History                  & \estci{+0.241}{+0.137}{+0.345} & \estci{-0.035}{-0.072}{+0.004} & \estci{-0.276}{-0.384}{-0.164} \\
    Medicine                 & \estci{+0.251}{+0.139}{+0.357} & \estci{+0.001}{-0.064}{+0.063} & \estci{-0.249}{-0.336}{-0.163} \\
    Biology                  & \estci{+0.344}{+0.254}{+0.435} & \estci{-0.049}{-0.113}{+0.011} & \estci{-0.394}{-0.484}{-0.300} \\
    \midrule
    All                      & \estci{+0.168}{+0.120}{+0.215} & \estci{-0.049}{-0.074}{-0.025} & \estci{-0.217}{-0.257}{-0.177} \\
    \bottomrule
  \end{tabular}

  \caption{
    \Ac{ob} \(F_1\) of the no-context adapters minus that of the natural adapters trained at each window.
    Rows are ordered by truncation rate, on one evaluation configuration (\num{2048}-token window, no repetition penalty), with paired bootstrap \qty{95}{\percent} \acsp{ci} over documents.
    The change is the gap at \num{2048} minus the gap at \num{1024}, which equals, with its sign reversed, what the natural adapters lost to truncation; \enquote{c.\ science} stands for computer science.
  }\label{tab:truncation}
\end{table*}

\subsection{The merge scaling error and prior work on scale}\label{app:scale}

At \(N = 32\) the error of \cref{sec:validity} entered each adapter's own update at one part in \num{1024} of its trained magnitude rather than one in \num{32}.
The same arithmetic sat in two further helpers, which formed the pairwise \ac{lerp} merges of our interference runs and the top-3 merges of an earlier \ac{ob} retrieval run, and we report neither.
In \textsc{Peft} 0.18.1 \citep{mangrulkar2022peft}, the \texttt{linear} combination multiplies both factors of each adapter by the square root of its weight times its scaling \(\alpha / r\) and gives the merged adapter a scaling of one, so with weights \(1 / N\) and a shared \(\alpha / r\) it forms exactly the update of \sqrtlerp{}, while concatenation keeps each adapter's own update at its weight and adds no cross terms; neither reproduces our scaling.
\Citet{back2026understanding} find that concatenating three unscaled document adapters collapses under an \(N\)-fold perturbation and that rescaling each factor by \(1 / \sqrt{N}\) recovers most of the gap to linear averaging, LoRA-\textsc{Lego} \citep{zhao2025merging} rescales merged adapters for the decay in norm that averaging causes, and LoRA Soups \citep{prabhakar2025lora} writes linear merging with its coefficient on both factors.

\subsection{Evaluation footing}\label{app:footing}

Every table in this paper takes its denominators from one set of runs, scored under the configuration of its numerators.
The specialist denominators of our first cloze composition grid, taken from the archived bf16 runs, are \qty{7}{\percent} higher on average than those on the \num{4}-bit footing (\num{0.584} against \num{0.545}).
Of the \num{0.045} by which the uniform footing lowers the specialists' cloze gain over the base model, almost all lies on the specialists' side: they move by \num{-0.039} (\ci{-0.070}{-0.010}) and the base model by \num{+0.006} (\ci{-0.016}{+0.028}).
The footing's shifts in the composition grid come through the mean of ratios, since on the aggregate ratio it moves no composition \ac{ci} across zero, and the no-context and style-normalized gains move by amounts not distinguishable from zero.
The archived bf16 evaluation did not itself repeat exactly: two runs of the base model under the same code, configuration, and items agree on about five generations in six and on every generation in only a third of the documents, although their mean factual recall barely differs, and the records establish no cause; the \num{4}-bit runs, by contrast, agree on every generation where they overlap (\cref{app:repeats}).

\subsection{What repeats exactly}\label{app:repeats}

Where the runs on the \num{4}-bit cloze footing overlap, across five runs and four versions of the runner, they agree string for string on all \num{1393} generations compared (\num{518} distinct generations, most of them compared in more than one pair of runs), including the retrieval run's cells for each query's own adapter, which reproduce the specialist denominators although that run loads and merges other adapters on the same base between them, as the \num{150} \ac{ob} specialist scores repeat under one configuration.
That is the standard against which a reproduction can be judged, and its scope is narrower than determinism in general: every one of these runs used one cluster queue, the two archived bf16 base-model cloze runs, from the same code and items, agree on only \num{883} of \num{1050} generations for reasons their records do not establish, and the archived \num{4}-bit composition numerators match the cells they share with the retrieval run on \num{64} of \num{66} evaluations at the six decimals the archive stores.

\section{Further related work}\label{app:related}

The closest work in framing is that of \citet{back2026understanding}, who train adapters per document and per chunk, route among them by text embedding and by token-level routers, merge the three highest-ranked, and score answers with a language-model judge, both \ac{cb} and with retrieved text in context.
They measure how much a module stores and how routing and merging lose it, whereas we ask what the standard \ac{ob} probe measures and whether the geometry of the weights indexes content; their retrieval matches query text to documents, ours adapters to adapters.
With the ground-truth modules merged by \ac{ties}, their performance falls as the number merged grows from one to five, which is consistent with our cloze grid (\cref{sec:cloze-grid}), although probe, merge operator and metric all differ.

Two re-evaluations of parametric \ac{rag} add details that \cref{sec:related} omits.
\Citet{tang2025understanding} find that token \(F_1\) had rewarded the short, template-like answers of the parametric variant, and \citet{rakotonirina2026where}, whose merged per-document adapters collapse with a single distractor document, find four merges within about one point of each other at five documents: averaging, concatenation, \ac{ties}, and DARE \citep{yu2024language}, which drops a random fraction of each update's entries and rescales the~rest.

\Citet{gekhman2024does} find that fine-tuning acquires new factual knowledge slowly and that doing so can increase hallucination.
Per-entity adapters trained on a knowledge graph store facts that, \ac{cb}, only the entity's own adapter recovers, and neither text embedding nor weight-space distance finds that adapter in a bank that holds it \citep{pulici2026storage}; our bank never holds the query's own adapter and asks instead whether a neighbour's can stand in, which on cloze it cannot detectably do (\cref{sec:retrieval}), and on \ac{ob} every adapter stands in about equally well (\cref{sec:transfer}).

Document adapters are close to the most favourable case for the interference that merging methods address, since their updates are nearly orthogonal layer by layer (\cref{sec:geometry}), and neither semantic nor weight-space distance anticipates what a \ac{ties} merge of two adapters costs (\cref{app:interference}).
KnOTS \citep{stoica2025model} and Core Space merging \citep{panariello2025accurate} merge full updates into an adapter whose rank may exceed that of its parts, so neither offers a matched-rank comparison for our merges, and \cref{app:scale} sets the \(1 / N^2\) scale of factor averaging against the rescaling of \citet{back2026understanding} and LoRA-\textsc{Lego} \citep{zhao2025merging}.

\end{document}